\documentclass[aps,pra,reprint,longbibliography]{revtex4-1}

\usepackage{array}
\usepackage{graphicx}
\usepackage{amsmath}
\usepackage{amssymb}
\usepackage{textcomp}

\begin{document}
	
\title[]{Superconducting Optoelectronic Neurons \uppercase\expandafter{\romannumeral 5 \relax}: Networks and Scaling}
	
\author{Jeffrey M. Shainline, Jeff Chiles, Sonia M. Buckley, Adam N. McCaughan, Richard P. Mirin, and Sae Woo Nam}
\affiliation{National Institute of Standards and Technology, 325 Broadway, Boulder, CO, 80305}		
	
\date{\today}
	
\begin{abstract}
Networks of superconducting optoelectronic neurons are investigated for their near-term technological potential and long-term physical limitations. Networks with short average path length, high clustering coefficient, and power-law degree distribution are designed using a growth model that assigns connections between new and existing nodes based on spatial distance as well as degree of existing nodes. The network construction algorithm is scalable to arbitrary levels of network hierarchy and achieves systems with fractal spatial properties and efficient wiring. By modeling the physical size of superconducting optoelectronic neurons, we calculate the area of these networks. A system with 8100 neurons and 330,430 total synapses will fit on a 1\,cm $\times$ 1\,cm die. Systems of millions of neurons with hundreds of millions of synapses will fit on a 300\,mm wafer. For multi-wafer assemblies, communication at light speed enables a neuronal pool the size of a large data center ($10^5$\,m$^2$) comprising 100 trillion neurons with coherent oscillations at 1\,MHz. Assuming a power law frequency distribution, as is necessary for self-organized criticality, we calculate the power consumption of the networks. We find the use of single photons for communication and superconducting circuits for computation leads to power density low enough to be cooled by liquid $^4$He for networks of any scale.
\end{abstract}
	
	
\maketitle

	
\section{\label{sec:introduction}Introduction}
Neural computation depends critically on the structure of the network \cite{spto2000}. Whether the function of the network is for control, sensory processing, or cognition, efficient information processing requires the network architecture to possess certain features. For example, in visual processing, local clusters of neurons must be differentiated to independently code for certain stimuli, yet the information from many such clusters must be combined at a larger scale to identify groupings of features and trends across a visual field \cite{haah2017}. Such differentiated local processing combined with broad integration repeats at multiple levels of hierarchy \cite{stsa2000,budr2004}. This hierarchical architecture leads to systems with fractal properties in space and time \cite{bu2006,busp2009}. Such an architecture balances differentiated, local information processing with efficient integration of information across the system \cite{to2004}. 

It has been proposed \cite{shbu2017,sh2018a} that combining the strengths of light for communication and superconducting electronics for efficient computation may offer a route to large-scale neural systems. Devices that perform the necessary neural operations have been described in Refs.\,\onlinecite{sh2018b,sh2018c,sh2018d}. In the present work, we begin to explore networks in which these superconducting optoelectronic neurons can be connected to achieve the desired fractal network architectures.

Useful superconducting optoelectronic networks (SOENs) must meet several criteria: 1) Physical instantiations utilizing superconducting electronics and dielectric waveguides must accomplish the networks in a manner that can be straightforwardly fabricated with conventional lithographic techniques; 2) The networks must achieve a hierarchical architecture that can be fabricated from the scale of a single die up to a 300 mm wafer; 3) For information integration and system synchronization, systems at the die scale must contain hub nodes with thousands of synaptic connections, and at the scale of a wafer, which we would like to serve as a region of cortex, high-degree nodes with tens of thousands of edges must be possible; 4) Considering these systems as modules, we extrapolate to neural systems at very large scales. It must be possible to connect modules with dense local clustering into systems of billions of neurons for human-scale cognition with the investment of reasonable materials and capital; 5) The power density of these networks must be low enough for cooling with $^4$He at 4.2\,K to be utilized. In this work, we present designs of SOENs meeting all these criteria.
	
\section{\label{sec:conceptualOverview}Conceptual Overview}
Optical communication between neurons provides three major strengths. First, because photons are uncharged and massless, optical interconnects have no capacitance or inductance, enabling massive fanout. Nodes with very high degree can be achieved without the need for multiplexed communication lines and signal arbitration \cite{lide2015}. Second, it is possible to send and receive single-photon signals, leading to communication with high energy efficiency \cite{mave2013}. Third, because light travels at the highest velocity possible in the universe, systems signaling with light can integrate the largest area of neurons with coherent oscillations. Yet devices based on optical signals have an important disadvantage: the device size is difficult to shrink below the wavelength of light. While superconducting optoelectronic neurons can make many connections, operate with high energy efficiency, and integrate a large area with coherent oscillations, the total number of neurons that can cooperate coherently depends on the total area of the system (limited by the speed of light) divided by the area of the individual neurons (limited by the wavelength of light). To begin to assess the potential of SOENs for near-term technological applications, we must analyze the types of systems that can be achieved on a single die, such as 1\,cm $\times$ 1\,cm. To assess the ultimate scaling potential of these networks when limited by physics, we must analyze what can be achieved within an area limited by the distance light can travel in the period of a network oscillation. We consider both of these scales in this paper.

Whether for near-term technological applications or long-term massive cognitive systems, the networks we wish to employ are likely to share several characteristics \cite{busp2009}. For cognitive systems large and small, differentiated processing balanced with information integration across spatial and temporal scales is crucial for performance \cite{tosp2003,to2004,to2008,bu2006,sp2010,base2011,haah2017} (see also discussion in Ref.\,\onlinecite{sh2018e}). Several network theory metrics can be employed to assess the fitness of a network for neural computing. We focus on three metrics relevant to differentiated processing and information integration. 

The first metric is clustering, which quantifies the prevalence of triangles in the network. Triangles refer to groups of three nodes all connected by edges. High clustering enables information to be shared locally. To achieve differentiated information processing, we seek networks with a high degree of clustering compared to a random network with the same number of edges on the same vertex set. 

The second network metric is average path length. For a given network, we can calculate the shortest distance from every node to every other node. Averaging this distance over all pairs of nodes yields the average path length. To achieve information integration across spatial scales, the average path length should be nearly as small as the corresponding random network. High clustering with short path length characterize a small-world network \cite{wast1998}. 

The third network metric is the degree distribution. The degree of a node refers to the number of edges it forms with other nodes. In the present work, we consider two degree distributions: power law and delta function. For information integration across spatial scales with efficient wiring \cite{busp2012}, the degree distribution of the nodes in the network may follow a power law, thereby achieving a scale-free network \cite{baal1999}. Small-world, power-law networks have efficient communication, a balance of differentiation and integration, fractal properties of self-similarity across spatial scales, and are observed in many natural settings, including the networks of the brain \cite{egch2005}. While power-law scaling has been observed in functional networks of the brain \cite{egch2005}, it is not clear if a power law degree distribution applies to neurons and their synapses in cortex \cite{spzw2004}. Further anatomical data is required to elucidate the degree distribution of neurons in cortex \cite{spto2005}. Therefore, we also consider structures closer to random networks, characterized by Gaussian degree distribution, which are optimized for short path lengths. Highly connected random networks with short path lengths are ideal for associative memories, such as hippocampus. We approximate the narrow Gaussian with a delta function, meaning all nodes of the network have the same degree. 


To achieve information integration across time, concepts of self-organized criticality \cite{bata1987,yara2017} are pertinent. In the temporal domain, differentiated local processing combined with large-scale information integration result in a power law frequency distribution of transient synchronized oscillations \cite{budr2004,bu2006}. This $1/f$ behavior leads to fractal use of time as well as space, and gives rise to the neuronal avalanches \cite{be2007} and criticality thought to be necessary for information integration and cognition \cite{be2007,kism2009,shya2009,rusp2011}. In the human brain, $1/f$ behavior is observed from 0.05\,Hz to 600\,Hz, spanning four orders of magnitude \cite{budr2004,bu2006}. For SOENs to achieve criticality, the devices and the network must support oscillations with frequencies obeying a power law distribution \cite{bu2006} over several orders of magnitude. As we have argued in Refs.\,\onlinecite{sh2018b,sh2018c,sh2018d}, SOENs will support oscillations from 1\,Hz up to at least 20\,MHz, thereby spanning seven orders of magnitude of temporal scale. 

The objectives of this paper are to 1) design networks with high clustering, short average path length, and power-law degree distribution that are suitable for differentiated processing combined with hierarchical information integration, 2) to approximate the area and power consumption of such networks when implemented with superconducting optoelectronic hardware, and 3) to anticipate the scaling of these networks when limited by physics. Using the tools of network theory \cite{eskn2015} and the principles gleaned from small-world and scale-free networks, we can design networks with the desired architecture. We focus here on micro-architecture at the level of neuron connectivity with the goal of identifying scaling laws for systems achievable in the near term. Each neuron will produce optical signals \cite{buch2017,sh2018d,bu2018}, and they will be routed via integrated-photonic waveguides \cite{chbu2017,sami2017} to many superconducting synaptic connections \cite{sh2018b,sh2018c}. To develop a correspondence between the adjacency matrix of the network \cite{eskn2015} and the physical hardware that will perform the neural operations \cite{sh2018b,sh2018c,sh2018d}, we model the spatial extent of neurons and construct a routing scheme to calculate the area of waveguide interconnects (analogous to white matter in the brain). In Sec.\,\ref{sec:networkConstruction} we discuss an algorithm which generates networks with the desired structural properties, and in Sec.\,\ref{sec:physicalInstantiation} we discuss the physical instantiation of the adjacency matrix using superconducting optoelectronic neurons. In Sec.\,\ref{sec:discussion} we consider near-term applications before transitioning to speculation about the future of this technology.
	
\section{\label{sec:networkConstruction}Network Construction}
\begin{figure} 
	\centerline{\includegraphics[width=8.6cm]{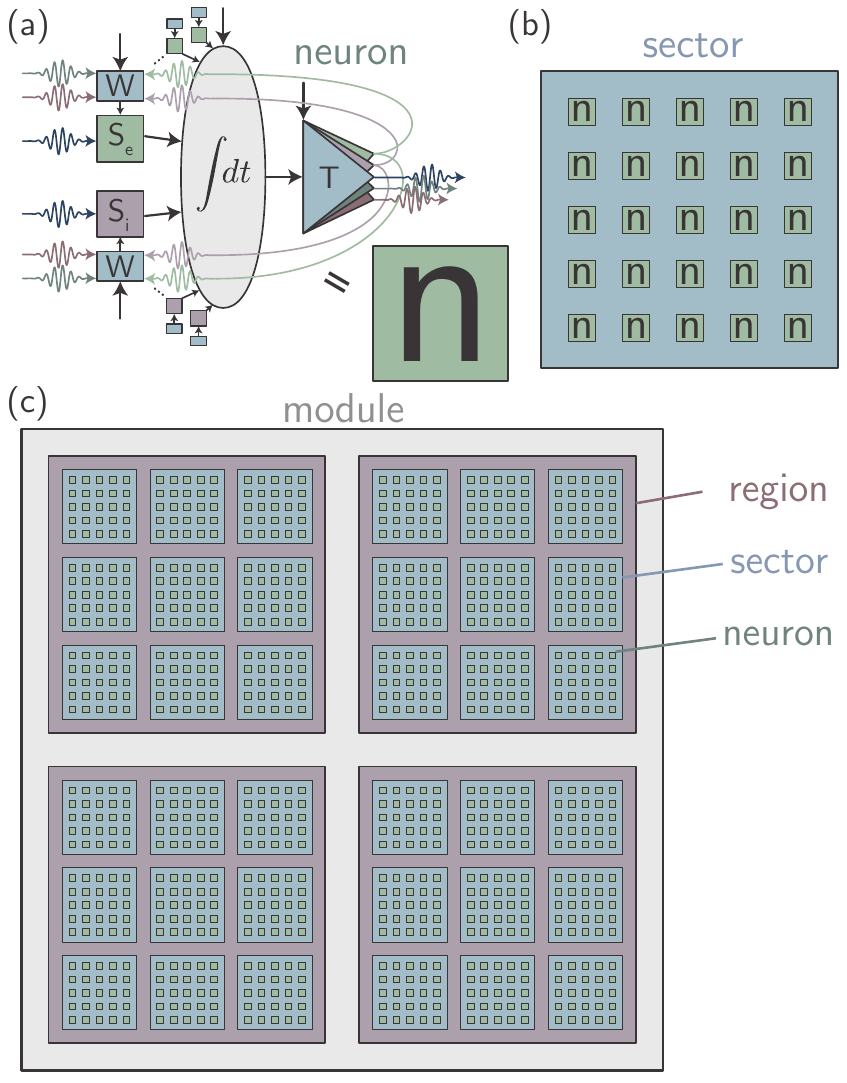}}
	\caption{\label{fig:networks_schematic}Schematic of networks under consideration. (a) Relating the neuron schematic used throughout this series of papers to the symbol used here. (b) Diagram of a sector of neurons. (c) Diagram of a module comprising multiple sectors and regions.}
\end{figure}

As mentioned in Sec.\,\ref{sec:conceptualOverview}, we design our networks based on three structural features. First, the network must have a high degree of clustering. Second, the shortest path between any two nodes averaged over all nodes in the network must be small. Third, the degree distribution of the nodes in the network will follow a power law. To achieve these goals while setting ourselves up for straightforward, modular fabrication, we construct our adjacency matrix in a hierarchical manner, as shown schematically in Fig.\,\ref{fig:networks_schematic}. At the smallest network scale, neurons are tiled in a grid to form a local sector (Fig.\,\ref{fig:networks_schematic}(b)). Sectors can then be tiled to form regions. Regions can then be tiled to form modules. In the present work, we consider these three levels of hierarchy, but the algorithm for creating the adjacency matrix can be repeated indefinitely. 
Networks constructed in accordance with the schematic of Fig.\,\ref{fig:networks_schematic}(c) with dense intra-sector connectivity, few connections between sectors within a region, and sparse connectivity across a module are modeled after the horizontal connections in mini-columns and columns within cortex \cite{mo1997,spto2000,brto2006,haah2017}. Feed-forward, vertical connections are likely to be employed by stacking such networks, either through sequential processing or multi-die assemblies.

To gauge clustering, average shortest path, and degree distribution in a network, it is often advantageous to compare to a random graph with the same number of edges constructed on the same vertex set. A regular graph \cite{wast1998,eskn2015} with few-neighbor edges will have a high clustering coefficient, but such a network will not achieve the short path length required of small-world architectures. One could employ the procedure of Ref.\,\onlinecite{wast1998} and replace edges between neighbors with distant connections at random to achieve a small world, but this procedure does not result in networks with a power-law degree distribution. Power law networks can achieve information integration across large regions of space with fewer nodes of high degree. This reduces the total area due to wiring \cite{bagr2010} (manifest as dielectric waveguides in the networks under consideration). In the present work, we consider power-law-distributed networks, and we compare to networks of nodes with uniform degree in calculations of area. Such delta function degree distribution may be necessary for certain cortical operations, such as the associative memory performed by hippocampus, which has such an architecture \cite{bu2006}.

To achieve high clustering while maintaining average path length close to that of a random network and introducing nodes with high degree, we utilize an algorithm intended to emulate the growth of naturally occurring networks. Such networks expand over time as new nodes are added, and new nodes have a higher probability of forming edges with existing nodes of high degree \cite{baal1999}. The growth algorithm utilized here is described in Appendix \ref{apx:growthAlgorithm}. The concept is that the adjacency matrix of a local sector of neurons is formed by adding neurons one by one to the pool, establishing edges based on spatial location as well as the degree of the existing nodes. Such an algorithm introduces winner-take-more development that leads to a few nodes with very high degree, thereby extending the degree distribution to larger values. The next level of hierarchy can be generated by tiling the adjacency matrix of the sector along the diagonal of a larger adjacency matrix representing the region, and forming inter-sector connections in a manner that again depends on space and node degree. Networks of increasing scale can be achieved by repeating this procedure, thereby achieving the fractal, power-law degree distribution conducive to efficient information integration across the physical substrate of the network. The spatial dependence gives high clustering, which leads to functional differentiation. Reciprocal connection rules contribute to clustering as well as reentrant connections \cite{spto2000}, which give rise to transient temporal correlations that are necessary for information integration across time. The degree dependence of the growth algorithm gives rise to the long tail of the degree distribution, including nodes with high degree, which are advantageous for information integration across space.

We compare a network generated with the growth algorithm to two other networks: a random network on the same vertex set, and a network that uses the growth algorithm to form local sectors, but establishes long-range connections between sectors and between regions in a manner that depends only on the distance. We refer to this as the partial growth network. The sectors of our calculations comprise a 9 $\times$ 9 grid of neurons. The regions comprise a 5 $\times$ 5 grid of sectors. The module comprises a 2 $\times$ 2 grid of regions. The total number of nodes is 8100, and the total number of edges in the growth and random network is 330,430. The partial growth network has the same number of nodes, and 304,365 edges. The data that summarize the small-world properties of the networks are shown in Table\,\ref{tab:smallWorld}. For comparison, we introduce the small-world index (SWI) \cite{hugu2008} given by  $\mathrm{SWI} = \frac{\bar{C}/\bar{L}}{\bar{C}_{\mathrm{r}}/\bar{L}_{\mathrm{r}}}$, where $\bar{C}$ is the average clustering coefficient, $\bar{L}$ is the average shortest path, and the subscript, $\mathrm{r}$, refers to a random graph. The clustering coefficient is discussed in Appendix \ref{apx:clusteringCoefficient}. 
\begin{table}
\begin{tabular}{|c|c|c|c|}
\hline
 & CC & APL & SWI \\ \hline
Random  & 0.005 & 2.81 & 1 \\ \hline
Partial growth & 0.065 & 2.94 & 12.3 \\ \hline
Growth  & 0.215 & 3.01 & 40.0 \\ \hline
\end{tabular}
\caption{\label{tab:smallWorld}Small-world properties of the three networks under consideration. The clustering coefficient (Appendix \ref{apx:clusteringCoefficient}) is denoted by CC. The average shortest path length is denoted by APL. The small-world index is denoted by SWI.}
\end{table}
We find that while the partial growth network achieves a slightly smaller path length than the growth network, the path length of the growth network is still very close to that of the random network. The significantly higher clustering coefficient of the growth network more than compensates, indicating that the growth algorithm is satisfactory for producing networks with small-world attributes.

\begin{figure} 
	\centerline{\includegraphics[width=8.6cm]{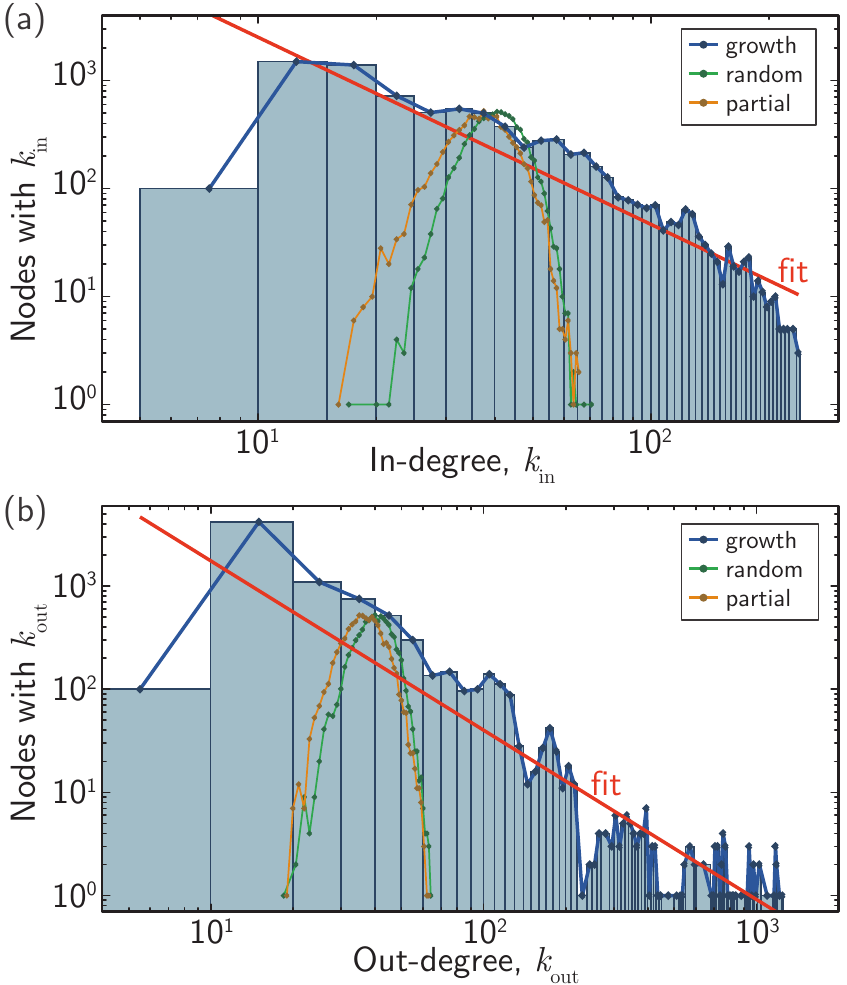}}
	\caption{\label{fig:networks_histograms}Histograms of growth, partial growth, and random networks. (a) In-degree. (b) Out-degree. The red lines are fits to the degree distributions of the network generated with the growth algorithm.}
\end{figure}
The degree distributions of the three networks are shown in Fig.\,\ref{fig:networks_histograms}. Here the significance of the growth model is prominent. In  Fig.\,\ref{fig:networks_histograms}(a) the in-degree of the networks is shown, and in Fig.\,\ref{fig:networks_histograms}(b) the out-degree is shown. For both in- and out-degree, the distribution of the growth network has been fit to a power law of the form $N(k) = B k^{-\gamma}$, where $k$ is the node degree. While the random and partial-growth networks display Gaussian degree distribution, the growth network fits well to a power law with $\gamma = 1.73$ for the in-degree and $\gamma = 1.64$ for the out-degree. The growth network contains many nodes with hundreds of edges, and several nodes with more than 1000 out-directed edges. Further comparison of the three networks is presented in Appendix \ref{apx:networkComparisons}, and general characteristics of networks characterized by power law degree distribution are considered in Appendix \ref{apx:spatialPowerLawRamifications}.  

The high clustering, short path length, and power-law degree distribution of the growth network meet our requirements for neural systems capable of control, sensory processing, and cognition. While these are some of the metrics by which we wish to gauge a network for dynamical neural systems, these metrics cannot be applied to feed-forward networks that are common for many machine-learning applications. To have high clustering, a graph must have many triangles. In graph theory, a triangle is a path with three edges that terminates where it begins (cycle of length three) \cite{eskn2015}. Feed-forward networks have no triangles because all out-directed connections go to the next layer, and no connections return from that layer. Thus, feed-forward networks have a clustering coefficient of zero. Because a node in one layer cannot reach another node in the same layer, the average shortest path length is infinity. Feed-forward networks with all-to-all connectivity between layers have a narrow degree-distribution. Thus, feed-forward networks are qualitatively different from the highly recurrent, small-world, power-law-distributed networks explored here for dynamical neural systems. Nevertheless, convolutional neural networks \cite{lebe2015} demonstrate features of differentiated processing combined with information integration. 

It is important to disambiguate two senses in which a network may have fractal spatial properties. We have been discussing the power law degree distribution, which relates to a network that is scale-free in terms of connectivity. We can also consider the spatial scales across which connections are made. Rentian scaling quantifies fractal properties related to the number of connections made through various partitions of the network. This scaling relates the number of nodes within a topological partition, $n$, to the number of edges crossing the boundary of that partition, $e$ \cite{bagr2010}. If the relation between $n$ and $e$ follows the form $n\propto e^p$, the network shows fractal topology, and the Rent exponent, $p_T$, is given by $p_T = \mathrm{log}_{10}(n)/\mathrm{log}_{10}(e)$. The Rent exponent is related to the topological dimension, $D_T$, by $p_T \ge 1-\frac{1}{D_T}$. Consideration of Rentian scaling and topological dimension provide a means to assess a network's connectivity across various levels of hierarchy. 

In the present context, Rentian analysis provides one way to assess the potential for information integration across the physical space of the network. The larger the topological dimension, the more access nodes will have to distant members of the network. For the hierarchical network designed with the growth algorithm, we can approximate $p_T$ by considering the number of nodes and edges at each of the three levels of hierarchy. Averaging over the neurons in a local sector, we find there are 17.1 edges into each neuron from other neurons within the sector. Considering a region, there are 17.8 edges into each neuron from other neurons within the region, but not in the same sector. And at the scale of the module there are 17.6 edges into each neuron from neurons in other regions. Based on this analysis, we calculate a Rentian exponent of $p_T = 0.99$. This gives a topological dimension of $D_T \rightarrow \infty$. This implies the network can integrate information up to infinite levels of hierarchy, but this is an approximation, as the network only comprises three levels of hierarchy. Networks with large topological dimension support self-organized criticality \cite{be2007,rusp2011}, which is advantageous for information integration in neural systems \cite{be2007,kism2009,shya2009,ch2010,rusp2011}. In networks from C. elegans to the human brain as well as in VLSI wiring, the Rentian exponent is closer to 0.75 \cite{bagr2010}. This analysis indicates that the network considered here, has slightly more long-range connectivity than is necessary. Area estimates based on this network (Sec.\,\ref{sec:physicalInstantiation}) will be slightly larger than the networks ultimately used in practice. The Rentian exponent can be adjusted within the grown algorithm by adjusting the number of connections made to higher levels of hierarchy (Appendix \ref{apx:growthAlgorithm}).  

We have constructed an adjacency matrix satisfying criteria for differentiated local processing and information integration. We must now assess the physical properties of this adjacency matrix when manifest as a superconducting optoelectronic neural system.
	
\section{\label{sec:physicalInstantiation}Physical Instantiation}
We have described an algorithm for generating adjacency matrices corresponding to networks with dense clustering, short path lengths, and power-law degree distributions. The task is now to calculate the physical aspects of a SOEN including size and power dissipation. We first introduce a plausible routing scenario for connecting the neurons in the network. 

\subsection{Passive waveguide routing}
At this stage in the development of the superconducting optoelectronic hardware platform, there is significant uncertainty regarding fabrication process and details of device implementation. Our objective here is to obtain a rough estimate of the size of networks based on a plausible routing and device layout scenario. To obtain this estimate, we must first consider how the neurons in the network will be connected via passive dielectric waveguides. 

To achieve the massive connectivity required of a neural system, each neuron must send its optical signals to many different destinations. To achieve this with optical signals in a controlled manner, multiple planes of dielectric waveguides \cite{chbu2017,sami2017} must be employed to keep waveguide crossing losses as low as possible and to reduce the area occupied by passive routing. We anticipate employing waveguide planes in pairs. Within each pair, one plane runs predominantly north-south, the other predominantly east-west. The vertical gap between waveguides in a pair is smaller than between the top waveguide in one pair and the bottom waveguide of the subsequent pair. The lowest pairs of waveguides will be used for the most local connections, with higher pairs for successively longer distances. Materials with lower index contrast may be used on higher planes to minimize loss in long-distance connections. In the context of the network discussed in Sec.\,\ref{sec:networkConstruction}, it is natural to use one pair of waveguide planes for intra-sector connections, a second pair for intra-regional connections, and a third pair for intra-modular connections.

\begin{figure}[h]
	\centerline{\includegraphics[width=8.6cm]{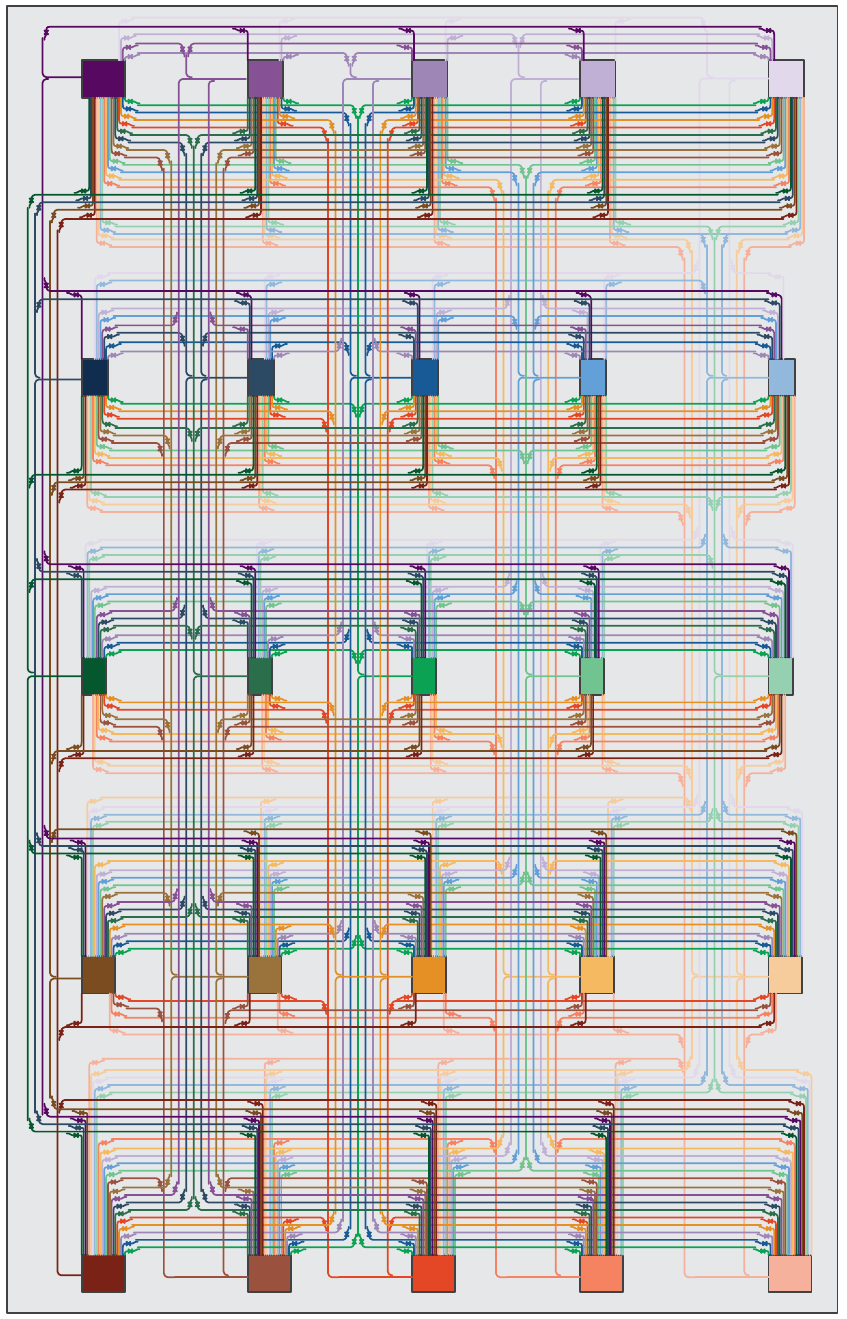}}
	\caption{\label{fig:networks_routingDiagram_full_allToAll}Row-column routing architecture in 5 $\times$ 5 sector. All-to-all connections.}
\end{figure}

\begin{figure}[h]
	\centerline{\includegraphics[width=8.6cm]{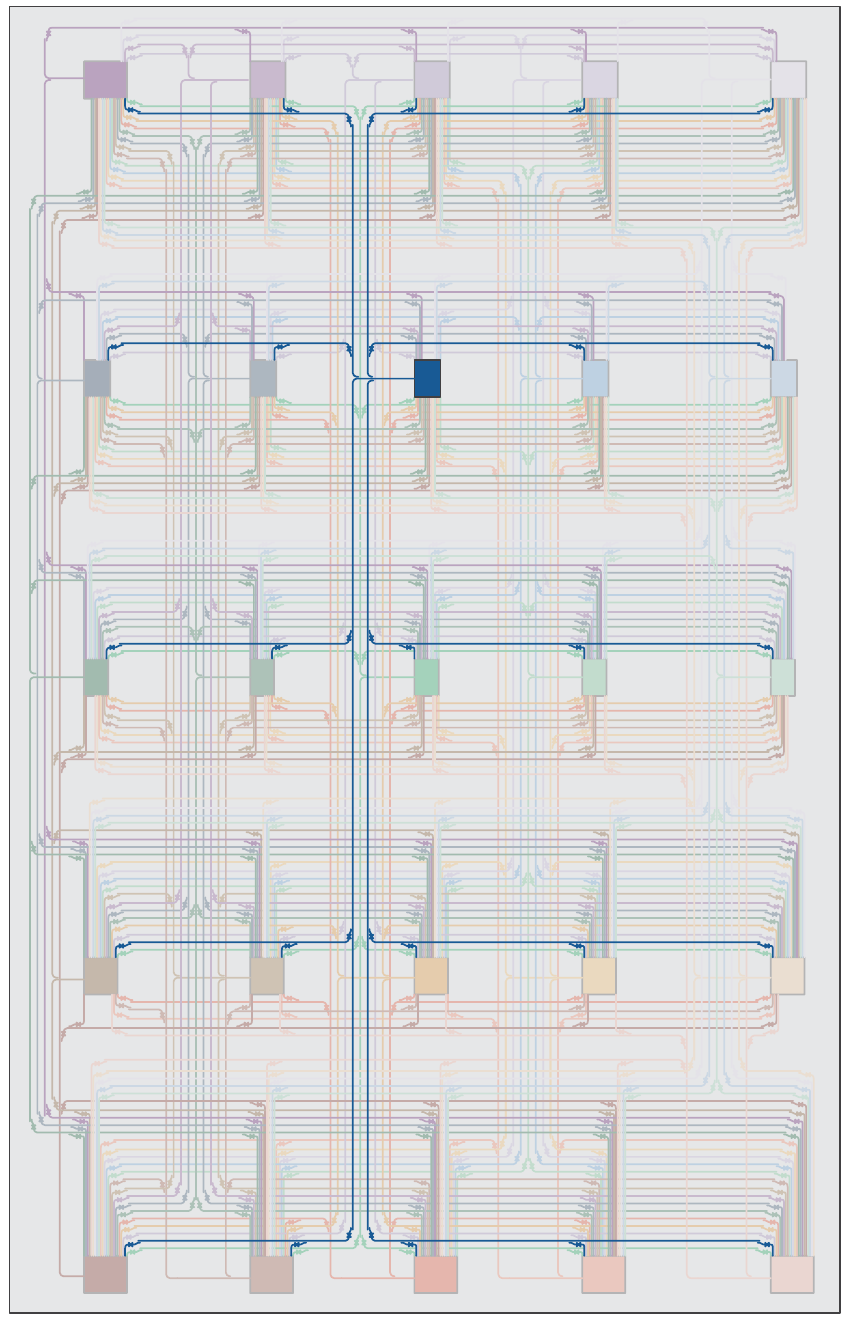}}
	\caption{\label{fig:networks_routingDiagram_full_oneToAll}Row-column routing architecture in 5 $\times$ 5 sector. The out-directed connections from a single node are highlighted.}
\end{figure}

In Fig.\,\ref{fig:networks_routingDiagram_full_allToAll} we show a row-column routing scheme that connects all neurons in a sector to all other neurons in that sector. This approach to routing results in rows and columns densely packed with communication lines, analogous to the white matter of axons in biological neural systems. The connections from a single neuron are highlighted in Fig.\,\ref{fig:networks_routingDiagram_full_oneToAll}, where inter-planar couplers are drawn as pairs of triangles. We utilize row-column routing to minimize the number of inter-planar couplers required. The rules governing the routing are described in Appendix \ref{apx:sectorRouting}. That appendix also discusses routing in feed-forward networks that are used in many machine learning neural networks.

While we have shown Figs.\,\ref{fig:networks_routingDiagram_full_allToAll} and \ref{fig:networks_routingDiagram_full_oneToAll} to depict neurons connected in a sector, the diagram can also be interpreted as showing sectors connected in a region or regions connected in a module. This fractal spatial property is a result of the hierarchical architecture we intend to employ for cognitive computing, and it is typical of networks characterized by high topological dimension. So far we have considered networks with three levels of hierarchy, but this fractal pattern can extend indefinitely. In the present context, we envision the row-column routing architecture extending beyond even the chip or wafer scale into multi-chip and multi-wafer assemblies with fiber optics forming dense white-matter connections between modules. Extension into the third dimension will be required. 

\subsection{Size of neurons} 
\begin{figure} 
	\centerline{\includegraphics[width=8.6cm]{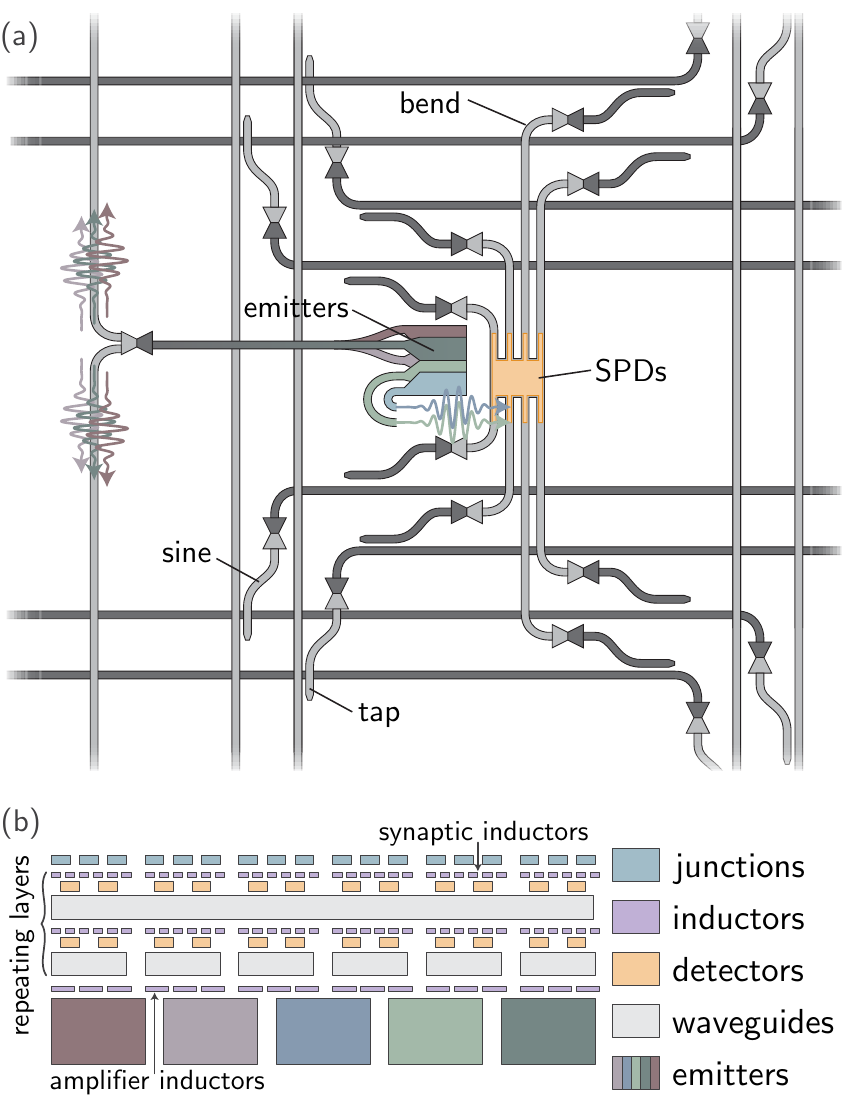}}
	\caption{\label{fig:networks_areaDiagram}Schematic of physical instantiation of neuron. (a) Pictoral motivation for area calculation in Appendix \ref{apx:sectorRouting}. Five emitters are shown producing photons for various synaptic operations \cite{sh2018c}. (b) Schematic of the fabrication layer stack. There is one layer of emitters at the bottom. Above the emitters are the inductors necessary to drive the emitters \cite{sh2018d}. Next are repeating layers of dielectric waveguides, superconducting detectors, and superconducting inductors. Josephson junctions are patterned above the waveguides and superconducting wires.}
\end{figure}

The size of the neurons described in Refs.\,\onlinecite{sh2018b} - \onlinecite{sh2018d}, when arranged in the networks described in Sec.\,\ref{sec:networkConstruction}, is calculated in Appendix \ref{apx:sectorRouting}. We use a simple spatial model of a superconducting optoelectronic neuron and routing waveguides shown in Fig.\,\ref{fig:networks_areaDiagram}(a). A schematic of how fabricated layers may be stacked is shown in Fig.\,\ref{fig:networks_areaDiagram}(b). 
Applying the analysis of Appendix \ref{apx:sectorRouting} to the network of 8100 neurons discussed in Sec.\,\ref{sec:networkConstruction}, we calculate that the network will fit on a die 1\,cm $\times$ 1\,cm using a pair of waveguide planes for intra-sector routing, another pair for inter-sector routing within each region, and a third pair for inter-regional connectivity. 

\begin{figure} 
	\centerline{\includegraphics[width=8.6cm]{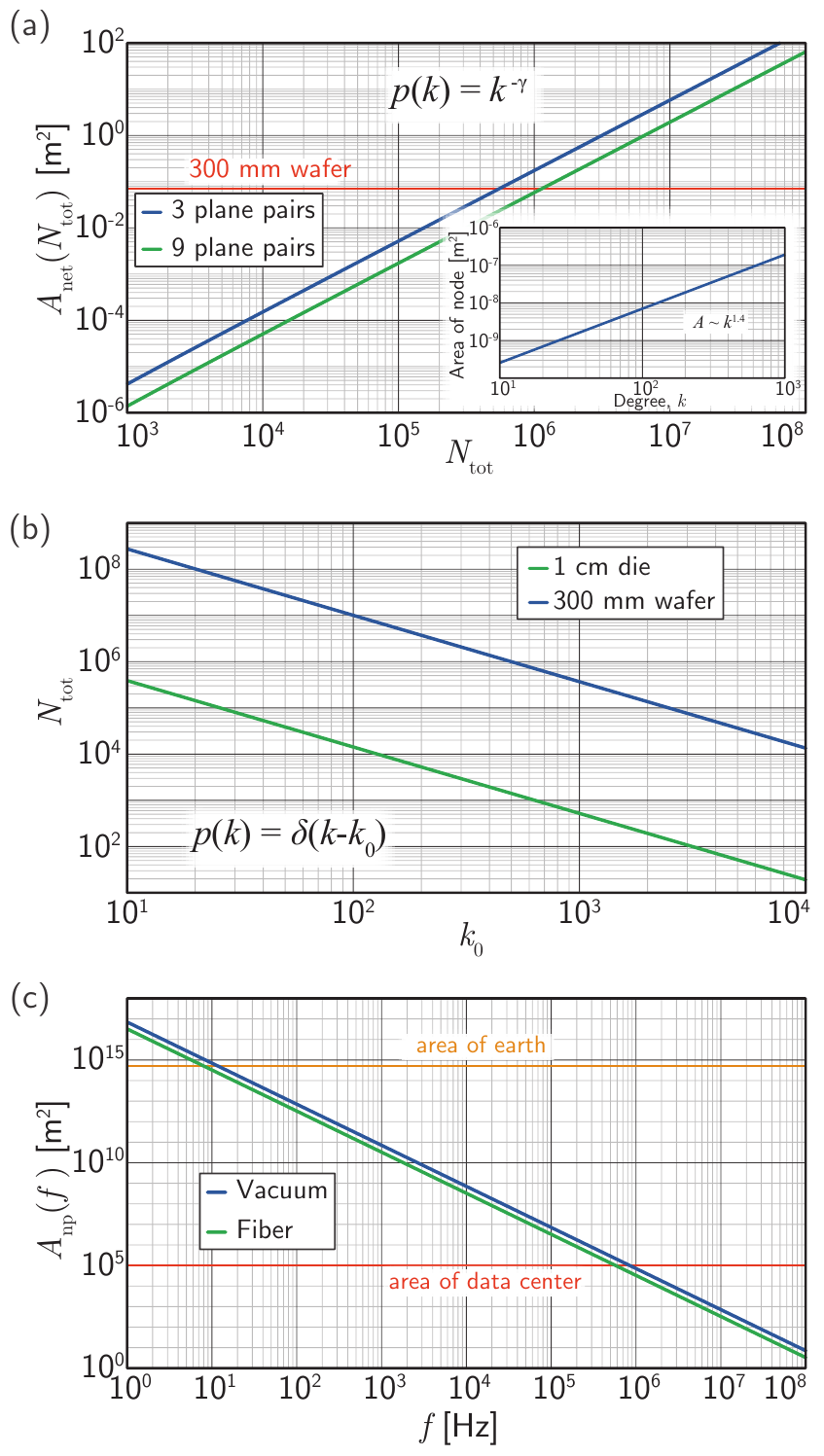}}
	\caption{\label{fig:networks_scaling}Scaling analysis. (a) Network with power-law degree distribution. Area as a function of total number of neurons for three and nine pairs of waveguide planes. The inset shows the area of a single node as a function of node degree when three routing planes are used. (b) Network with delta function degree distribution. The number of neurons that can fit on a 1\,cm $\times$ 1\,cm die and 300\,mm wafer as a function of node degree, $k_0$. (c) Area of the neuronal pool as a function of the frequency of oscillations for signals propagating in vacuum and fiber.}
\end{figure}

We would like to approximate the size of networks with more neurons. The model of Appendix \ref{apx:sectorRouting} informs us that the area of a neuron will follow a power law as a function of node degree. We fit the area per neuron calculated for the network of 8100 neurons to a power law and find an exponent of 1.4. We calculate the area of a network as a function of the total number of neurons in the network by extrapolating from the fit of the network of 8100 nodes and assuming a network with power law degree distribution following the trends of Appendix \ref{apx:spatialPowerLawRamifications}. The results of this calculation are shown in Fig.\,\ref{fig:networks_scaling}(a) for the cases of three pairs of waveguide planes and nine pairs of waveguide planes. In both cases, the degree distribution of the nodes takes the form of Eq. \ref{eq:pOfk} with $\gamma = 1.6$. We find that by utilizing nine pairs of waveguide planes, we can accommodate an integrated system of one million neurons on a 300\,mm wafer. This network would comprise over 200 million complex synapses, as described in Ref.\,\onlinecite{sh2018c}. Considering instead a network with delta function degree distribution (Fig.\,\ref{fig:networks_scaling}(b)), we find a network of 3000 neurons each with 300 synapses will occupy a 1\,cm $\times$ 1\,cm die. A 300\,mm wafer will support 40,000 neurons with 4000 connections. The number of synapses within a given area is similar for the power law degree distribution and the delta function degree distribution. The main difference is the power law degree distribution contains more neurons with fewer synapses and a few neurons with many synapses. Both network architectures are likely to be useful for certain types of information processing.

Appendix \ref{apx:sectorRouting} describes the calculation of the area of a neuron based on consideration of optical waveguides and single-photon detectors. From this area estimate, we can calculate the area available for the superconducting electronic components that contribute to the neuron \cite{sh2018b,sh2018c,sh2018d}, assuming they are patterned above the waveguides in repeating layer modules, as shown in Fig.\,\ref{fig:networks_areaDiagram}(b). The area available for superconducting inductors, mutual inductors, and Josephson junctions is given by the area of the node divided by the degree. This area is multiplied by the number of repeating layers utilized in the fabrication process. We find the area available for superconducting electronic components is at least 30\,\textmu m $\times$ 30\,\textmu m. A 1\,\textmu H inductor is on the larger end of what is likely to be utilized by the synaptic integration loop or synaptic storage loop \cite{sh2018b,sh2018c}. When constructed of a material such as MoSi with high kinetic inductance (180\,pH/$\square$) and patterned with 50\,nm lines, the area is 5\,\textmu m $\times$ 5\,\textmu m. This leaves plenty of area for multiple inductors and wiring. The mutual inductors required to couple the synaptic integration loop to the neuronal integration loop \cite{sh2018b} will be patterned with a different superconducting material separated from the high-kinetic inductance layer by an insulator. For these devices, 100\,pH to 1\,nH is desirable. Again, it is quite possible to fabricate a 1\,nH mutual inductor from two superconducting coils within this area. Each synapse also requires on the order of 10 Josephson junctions with $I_c$ near 40\,\textmu A. These devices are approximately 1\,\textmu m $\times$ 1\,\textmu m, and will easily fit in the available area. Finally, the largest components of the neurons are likely to be the amplifiers that drive the light emitters \cite{sh2018d} and the mutual inductor coupling the neuronal integration loop to the neuronal thresholding loop. These devices are not required at every synapse, but rather only at each neuron. Therefore, area is not likely to be a problem for these devices. Throughout this analysis we find passive photonic waveguides are likely to limit the area, and they are limited by the wavelength of light. Further considerations related to manufacturing are presented in Appendix \ref{apx:manufacturing}.

We have mentioned in this series of papers that neurons based on similar Josephson junction circuits, but using no optical components, could be created with much the same design as the neurons in Ref.\,\onlinecite{sh2018b} and synapses in Ref.\,\onlinecite{sh2018c}. The purely electrical neurons would have difficulty achieving the fanout necessary for the power law degree distribution discussed in the present work. Nevertheless, the electrical neurons could potentially achieve $k\approx$\,20 with a more compact footprint, lower power dissipation, and higher speed. It may be advantageous to construct a hybrid network with nodes from $k = 2$ to $k = 20$ being purely electrical neurons with flux quanta playing the role of photons. We consider a network with power law distribution wherein nodes with $k = 2$ to $k = 20$ are assumed to be electrical neurons, and nodes with $k > 20$ are assumed to be optical neruons. We perform an area calculation using modified forms of Eqns.\,\ref{eq:wCol} and \ref{eq:hRow} using physical dimensions appropriate to superconducting wiring rather than photonic waveguides. Whereas a network of just over one million optoelectronic neurons fit on a 300\,mm wafer, a network of 4 million total neurons can be achieved if we combine low-degree superconducting neurons with high-degree photonic neurons. This network would comprise over one billion synaptic connections. Such an architecture would extend the spatial power law distribution to lower degree while extending the temporal power law distribution to higher frequency. 
 
\subsection{The neuronal pool}
In addition to comparing to the size of a 300\,mm wafer we must also consider the largest SOEN that can be established. This limit is not technological or economic in nature, but rather it is a physical limitation set by the velocity of communication. 

Consider two neurons separated by a distance $d$. At time $t=t_0$, neuron one produces a pulse. At time $t= t_1$, neuron two produces a pulse. We ask whether the pulse generated by neuron one may have contributed to the pulse by neuron two. In particular, we are interested in whether neuron one may drive neuron two to synchronize. For this consideration, the time scale we consider is the inverse of the frequency of oscillation, $T = 1/f$. Thus, if 
\begin{equation}
\label{eq:neuronalPool_diameter}
d \le \frac{v}{f},
\end{equation} 
neuron one can induce neuron two to synchronize with oscillations at frequency $f$. For the present discussion, we take the value of $d$ that saturates the inequality of Eq.\,\ref{eq:neuronalPool_diameter} to define the diameter of the neuronal pool. This result of the size of the coherent neuronal pool decreasing as the inverse of the frequency of synchronized oscillations \cite{lued1997} is consistent with observations from neuroscience \cite{stra1999} where the slow conduction velocity of axons is a severe limitation to the size of a synchronized network \cite{budr2004,bu2006,buwa2012}. The conduction velocity can be increased in axons with larger diameter, but this introduces a size/speed trade-off. No such trade-off is present with optical signals. However, a size/power trade-off is present, because more compact waveguides can be implemented with high-index-contrast waveguides, but larger, low-index-contrast waveguides will have lower loss and therefore require fewer photons to be produced per synaptic connection. Based on Eq.\,\ref{eq:neuronalPool_diameter}, we see the area of the neuronal pool scales as $A_p \propto (v/f)^2$.

The number of neurons in the pool depends also on the size of the neurons. We suppose the neurons in a network can be characterized by an average width, $w$. If we assume a network is characterized by long-range connectivity in a plane (as is the case for mammalian cortex and as we expect from lithography on wafers), the area of a single neuron scales as $A_n \propto w^2$. For a given hardware platform sending signals at velocity $v$ between neurons with width $w$, the number of neurons that can be synchronized at frequency $f$ scales as $N_p \propto A_p/A_n \propto (v/fw)^2$. 

We wish to compare the number of neurons in the pool of two different hardware platforms oscillating at the same frequency, $f$. In particular, we wish to compare SOENs to the human brain. Denoting the two hardware platforms with superscripts, we find
\begin{equation}
\label{eq:neuronalPool_number}
\frac{N_p^{(\mathrm{soen})}}{N_p^{(\mathrm{brain})}} = \left(\frac{v^{(\mathrm{soen})}w^{(\mathrm{brain})}}{w^{(\mathrm{soen})}v^{(\mathrm{brain})}}\right)^2.
\end{equation}
As we have calculated above, a 300 mm wafer can support roughly $10^6$ superconducting optoelectronic neurons, giving $w^{(\mathrm{soen})} = 2.7\times10^{-4}$\,m. The area of the human cerebral cortex is 0.095\,m$^2$ \cite{scva2014}, and it contains $1.6\times10^{10}$ neurons \cite{azca2009,he2009}, giving $w^{(\mathrm{brain})} = 2.4\times 10^{-6}$\,m. A biological neuron is 100 times smaller than a superconducting optoelectronic neuron in width, and 10,000 times smaller in area. Yet the speed of signals in the cortex is roughly 2\,m/s \cite{edve2004}. Axons with larger diameter can propagate signals above 100\,m/s, but for the dense connectivity of the cortex, such large fibers cannot be supported. The speed of light is $3\times 10^8$\,m/s \cite{speedOfLight}. Thus, comparing SOENs to biological networks, we find $N_p^{(\mathrm{soen})}/N_p^{(\mathrm{brain})} \approx 10^{12}$. That is, the neuronal pool enabled by light-speed communication can contain a trillion times the number of neurons of the neuronal pool enabled by ionic signal propagation along biological axons. Signaling at the speed of light brings a tremendous advantage in this regard, and the advantage is even more significant if one considers networks spanning a volume rather than an area. 

It has been hypothesized that the conduction velocity of axons limits the size of biological neural systems, and that scaling of a single cognitive system much beyond the human cortex may not be possible with biological axon conduction velocities \cite{rido1994,bu2006}. Based on these simple scaling arguments, we expect the largest cognitive systems in the universe to communicate at the speed of light, enabling full utilization of each neuron's light cone. While other approaches to neural computing may utilize light for signaling, it is unlikely other devices will be significantly smaller than those presented here, as the optical wavelength imposes a practical spatial limit. While shorter wavelengths could be utilized, higher energy would be required per photon, and materials suitable to large-scale integration of light sources and waveguides below visible wavelengths are not amenable to construction of integrated systems. In Fig.\,\ref{fig:networks_scaling}(c) we show the area of the neuronal pool as a function of oscillation frequency $f$ for a general neural system communicating at the speed of light. When oscillating at 1\,MHz, an area the size of a large data center ($10^5$\,m$^2$) can be coherently synchronized. When oscillating at 10\,Hz (near the theta and beta bands of the human cortex), an area the size of the surface of the earth can by coherently synchronized. 

\subsection{Power consumption}
In addition to the size of a SOEN, we must also consider its power consumption, which depends on the energy per synapse event, the degree distribution of the neurons, and the firing frequency distribution of the neurons. Analysis of power consumption in the networks under consideration is detailed in Appendix \ref{apx:power}. From that analysis, we find if we assume total photon production efficiency of $\eta = 10^{-4}$ (see Ref.\,\onlinecite{sh2018d}), the network of 8100 neurons will dissipate 1\,mW of device power, a trivial load for a modern cryogenic system. Including the power dissipated to establish bias currents, total on-chip power is less than 10\,mW. The dominant power draw is, of course, the cryo cooler, and for this system it would require little more power than a household air conditioner (a few hundred watts).

For the system of one million neurons and 200 million synapses on a 300\,mm wafer, the total power dissipation would be approximately 1\,W. This is the cooling power of a common Gifford-McMahon cryocooler with base temperature of 4.2\,K. 

In addition to low total power, the power density of these systems is also quite small. For the case of the network on a wafer, the power density is 10\,W/m$^2$ if a poor total light-generation efficiency of $\eta = 10^{-4}$ is assumed. The heat produced can be straightforwardly removed, and immersion in a liquid helium bath will be excellent for this purpose \cite{ek2006}. This power density is far below the threshold for boiling liquid helium. Liquid helium is transparent at telecommunication wavelengths with an index of refraction very close to one. Thus, optical signals can be sent between neural modules in this environment without attenuation at the maximum speed allowed in the universe. Because the area of devices grows as a faster function of node degree than the power dissipation, cooling SOENs is possible in principle up to systems with size limited by the speed of light.

\section{\label{sec:discussion}Discussion}
We have argued that neural systems benefit from at least three network structural characteristics: clustering, short paths, and power-law scaling. We have utilized an algorithm to generate adjacency matrices with these characteristics that correspond to networks hierarchically arranged across spatial scales, and we have proposed that such networks could be implemented with superconducting optoelectronic hardware using 300\,mm lithographic technology. Networks with millions of neurons and hundreds of millions of synapses can be fabricated on a 300\,mm wafer. While Moore's Law has been sustained through feature size reduction, scaling of SOENs will be enabled by scaling the process up in terms of number of processed layers, lithographic field area, wafer size, and multi-wafer assemblies. Whereas economics has driven CMOS to bigger wafers, network scaling will drive SOENs to bigger wafers. Fortunately, feature sizes in all devices of this system are easily achievable with 190\,nm lithography.

We are interested in systems with near-term feasibility. Considering the aforementioned network of 8100 neurons and 330,430 synapses, the network occupies 1\,cm $\times$ 1\,cm when processed with 3 pairs of waveguiding planes. Operating at 20\,MHz and possibly far beyond, the dynamical state space of this system is extraordinarily complex and is likely to provide a fascinating experimental setting for many subjects including self-organized criticality \cite{be2007,kism2009,shya2009,ch2010,rusp2011} in an artificial network approximating cortex. Such a neural module is also likely to offer significant technological opportunities in applications such as visual and auditory sensory processing, language recognition, and mechanical control.  

When combined with an artificial retina based on superconducting detectors, a powerful artificial vision system will be achievable. A three-chip stack could accomplish this. One chip would serve as a retina, and the other two chips would serve as layers of visual cortex, arranged in a manner analogous to the columnar organization of the mammalian visual cortex \cite{mo1997}. Feed-forward communication is likely to make use of vertical optical links with light in a lower layer being sent vertically with a grating coupler and received by a single-photon detector at a neuron on a higher layer. Such a model, capturing the columnar, vertical flow of information between layers as well as the laminar, horizontal flow within layers \cite{lued1997,spto2000,brto2006}, has been shown to function well for object recognition in computer simulations \cite{haah2017}. Using visual systems as a testbed of the SOEN hardware is beneficial because photon detectors are native to the hardware, and also because the visual system is the most studied system of the mammalian cortex. Investigation of binding in vision \cite{ro1999,tr1999,woca1999,vala2001,enfr2001,haah2017} can inform other neural computing systems including motor control systems, language processing, or other cognitive architectures.

Drawing further analogy between the networks presented here and cortical mini-columns and columns \cite{mo1997,haah2017}, the entire module of 8100 neurons may be analogous to a basic anatomical component of neurocognitive function \cite{brto2006}. In the context of visual binding \cite{ro1999,tr1999,woca1999,vala2001,enfr2001,haah2017} that we discussed in Ref. \onlinecite{sh2018a}, we would expect the neurons in a sector to have similar tuning curves in response to visual stimulus, and therefore the population of each sector would code for certain aspects of a visual scene \cite{brto2006}. Integrating the information up the hierarchy, a module with 8100 neurons may perform well for detecting many aspects of visual stimulus, and therefore be capable of detecting a variety of objects \cite{haah2017}. Modules may be designated for different visual functions, such as attention to shape, color, or motion, and a multi-module assembly would comprise a complex, highly functional vision system.

By constructing complex networks of neurons with inhibition and dendritic processing, the hardware may prove suitable to investigate the role of synchrony in visual binding \cite{woca1999,enfr2001}. The neurons and networks described in this series of papers have been designed to include important anatomical features to enable synchrony, such as differentiated local processing combined with system-wide information integration \cite{to2004,bato2008,base2011}, oscillations at many frequencies enabled by inhibitory signals modulating the receptivity of dendritic spines \cite{vala2001,sase2001,enfr2001,budr2004,bu2006,robu2015,haah2015}, and synapses with diverse time constants employing event-based plasticity and metaplasticity to enable adaptability as well as long-term memory retention in environments with continually adapting stimulus \cite{bico1982,fudr2005,fuab2007,cobe2012}. Modules of visual cortex may be a promising first demonstration of the utility of this hardware for cognition \cite{br1995,toed1998,mc2000,to2004,bato2008,brme2010,base2011,fr2015,haah2017}.

Similar modules are also likely to tile well into larger networks. With a 10 $\times$ 10 grid of interconnected die, the system would contain as many neurons as a bumble bee, a creature observed to have advanced navigational skills \cite{chmi2014}; multimodal communication and learning \cite{alpe2016}; cognitive flexibility \cite{lope2017}; and emotions \cite{peba2016}. A 300 mm wafer would yield enough die for six of these small brains. Each system would comprise nearly a million neurons and over 33 million synapses with complex plasticity and metaplasticity \cite{sh2018c}. These neurons and synapses would operate up to thirty thousand times faster than any known living creature. Such a system would be readily accommodated by existing cryostats.  

If one chose to utilize an entire 300\,mm wafer for a single module designed with a growth algorithm, neurons with as many as ten thousand synapses could integrate the activity of networks with a million neurons. Such a module would be analogous to a module in cortex, except for the significant increase in speed. Considering such modules as the building blocks of larger systems, stacks of wafers could be assembled with free-space and fiber coupling between. We can consider an interconnected stack of wafers as a column. An arrangement of columns could be interconnected to contain tens of billions of neurons, matching the number in the human neocortex \cite{brsc1998,he2009}. A SOEN with the same number of neurons as human cortex would occupy a volume roughly one meter cubed. 

While it may be difficult to build systems larger than 10 billion neurons in the near term, such a system is not physically limited. Like the brain, such limits will be incurred due to the velocity of signal propagation. From Fig. \ref{fig:networks_scaling}(c) we know that networks as large as data centers can sustain coherent oscillations at 1\,MHz. Such a facility would house $10^8$ 300\,mm wafers if they were stacked 100 deep. This would result in 100 trillion neurons per data center across modules interconnected with another power law distribution. 

Networks need not oscillate at 1\,MHz, and if they supported system-wide activity at 1\,kHz\textemdash faster than any oscillation of the human brain\textemdash the neuronal pool could occupy a significant fraction of the earth's surface and employ quintillions of neurons. We do not wish to cover earth in such devices, but asteroids provide ample, uncontroversial real estate. The materials for this hardware are abundant on M-type and S-type asteroids \cite{mufo2017,astra,bu1999,shcl2010,necl2014}. It appears possible for an asteroid belt to form the nodes and light to form the edges of a solar-system-scale intelligent network. Asteroids can be separated by billions of meters, so light-speed communication delays may be several seconds or longer. For cognitive systems oscillating up to 20\,MHz, such delays would cause individual modules to operate as separate cognitive systems, much like a society of humans. 

\vspace{0.5em}
This is a contribution of NIST, an agency of the US government, not subject to copyright.
	
\newpage
\appendix

\section{\label{apx:growthAlgorithm}Network growth algorithm}
The growth algorithm utilized in this work to achieve networks of power law degree distribution proceeds as follows. The number of nodes and spatial configuration of the local sector is chosen (a square grid throughout this work). The spatially central node is assumed to emerge first, and initially it has no connections. The remaining nodes are introduced to the network in order of their Euclidean distance from the central node. The new node has a probability of forming a connection to each existing node in the network. The probability is calculated as 
\begin{equation}
\label{eq:pOfL}
p(L) = p_0(L_{\mathrm{min}}/L_{\mathrm{eff}})^\alpha,
\end{equation}
where the in-degree-dependent effective length is given by 
\begin{equation}
\label{eq:lEff}
L_{\mathrm{eff}} = L-(L-L_{\mathrm{min}})(k_{\mathrm{in}}/\lambda k_{\mathrm{in}}^{\mathrm{max}})^{\beta}.
\end{equation}
Here, $L$ is the Euclidean distance, $L_{\mathrm{min}}$ is the minimum spacing between nodes in the network, $k_{\mathrm{in}}^{\mathrm{max}}$ is the maximum in-degree possible given the number of nodes in the sector, and $\lambda$ is a scaling factor that can be utilized to obtain different numbers of edges, clustering, and degree distributions. The exponents, $\alpha$ and $\beta$, are parameters of the model that can be adjusted to modify the connectivity across spatial scales. Each time a new neuron is added to the local sector it has a probability of forming an out-directed connection with each node in the local sector, and each node in the local sector also has the same probability of forming an out-directed connection to the new node. This rule is included to increase the number of reciprocal connections in the network, and when similar reciprocal rules are applied on subsequent levels of hierarchy, many re-entrant connections emerge \cite{spto2000}. With these rules, the probability of a new neuron forming a connection to an existing neuron in the network depends on both the distance between the two neurons (based on their spatial origins) and the degree of the existing neurons (which may be thought of as the existing neuron's size or influence). The algorithm for constructing the adjacency matrix corresponding to the local sector begins to introduce the winner-take-more phenomenon that is necessary for obtaining nodes with high degree.

To expand the network to the next level of hierarchy, we could produce unique adjacency matrices for each sector of the region. Instead, motivated by computational efficiency and the desire to make economical use of reticles during fabrication, we initialize the adjacency matrix of the region by tiling the adjacency matrix of one sector along the diagonal. Then, to ensure unique local processing clusters and to continue the growth of the power-law degree distribution, we introduce a winner-take-more effect at this level of hierarchy as well. We allow only a certain number of the nodes with highest degree ($N_{\mathrm{win}}$) the opportunity to receive connections from nodes in different sectors. At this stage, the algorithm steps through each of the nodes in a sector, and gives each of them several chances to form a connection to the $N_{\mathrm{win}}$ nodes of highest total degree in each of the other sectors. The number of chances for each connection is given by $N(k) = N_{\mathrm{min}}-(N_{\mathrm{min}}-\xi N_s)((k-k_{\mathrm{min}})/(k_{\mathrm{max}}-k_{\mathrm{min}}))^{\delta}$, where $N_s$ is the number of nodes in the sector and $N_{\mathrm{min}}$ and $\xi$ are parameters that can be used to adjust the network properties. $k_{\mathrm{max}}$ and $k_{\mathrm{min}}$ are the maximum and minimum total degrees of the nodes in the initial local sector. The exponent $\delta$ is a parameter of the model that can be adjusted to modify the degree distribution. The probability that each connection will be made is again a function of space calculated with Eq. \ref{eq:pOfL}, but here the true Euclidean length (not degree-dependent) is used, and each neuron is assumed to be at the origin of its local sector.   

To further scale the network, the same winner-take-more approach taken to scale a sector to a region can be used to scale a region to a module. The parameters present in the model can be adjusted to achieve a range of adjacency matrices with various clustering coefficients, average shortest path lengths, and degree distributions. We have used $p_0 = 1$ for sector generation, $p_0 = 0.3$ for region and module generation, $\alpha = \beta = \delta = 1.5$, $\lambda = 0.45$, $N_{\mathrm{min}}=1$, $\xi=0.75$. When assigning connections between sectors to form regions, half of the neurons in each sector were given the opportunity to form more distant connections ($N_{\mathrm{win}} = 41$ out of 81 possible). When assigning connections between regions to form a module, only the top 2.5\% of neurons were given the opportunity to form more distant connections ($N_{\mathrm{win}} = 51$ out of 2025). 

While simpler and more mathematically elegant approaches to generating cortex-like adjacency matrices have been introduced \cite{sp2006}, we are intrigued by the fractal result of the naturally motivated growth algorithm. Other growth algorithms are likely to be able to achieve networks with similar properties as the principles of spatial and winner-take-more connection probability are generally effective at producing small-world, scale-free networks. The number of parameters used in the growth model is not parsimonious, but the number is suitable to comprise a genome for evolutionary optimization.

\section{\label{apx:clusteringCoefficient}Clustering coefficient for binary directed networks}
To measure clustering in the directed networks under consideration, we utilize the clustering coefficient for node $i$ introduced in Ref.\,\onlinecite{fa2007}:
\begin{equation}
\label{eq:cc_i}
C_i^D(\mathbf{A}) = \frac{[(\mathbf{A}+\mathbf{A}^T)^3]_{ii}}{2[d_i^{tot}(d_i^{tot}-1)-2d_i^{\leftrightarrow}]}.
\end{equation}
Here, $\mathbf{A}$ is the adjacency matrix, $d_i^{tot}$ is the sum of in-directed and out-directed connections, and $d_i^{\leftrightarrow}=(\mathbf{A}^2)_{ii}$ is the number of bilateral edges between node $i$ and its neighbors. The total clustering coefficient for the network is the sum of clustering coefficients of all nodes divided by the number of nodes in the network. Throughout the present work, we are interested in basic network structure, so we consider binary directed networks and neglect the effect synaptic weights may have on the analysis. The clustering coefficient given by Eq.\,\ref{eq:cc_i} is the ratio of the number of directed triangles formed by node $i$ to the number of directed triangles node $i$ could form given the degree of node $i$.

\section{\label{apx:networkComparisons}Network comparisons}
\begin{figure} 
	\centerline{\includegraphics[width=8.6cm]{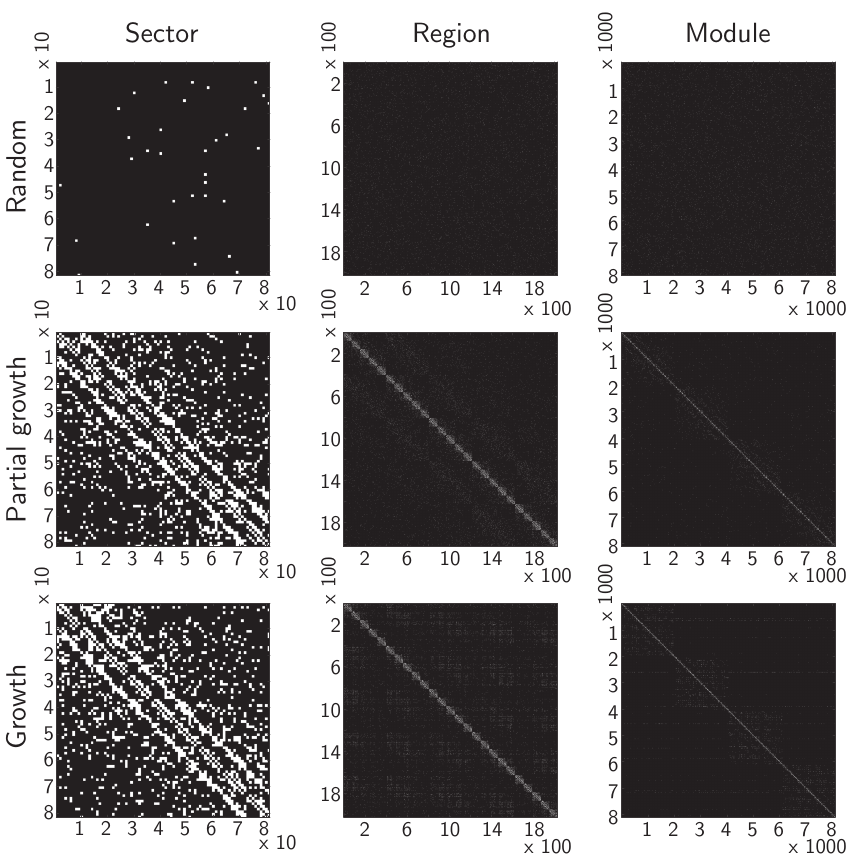}}
	\caption{\label{fig:networks_adjacencyMatrices}Adjacency matrices of random network, partial growth model, and growth algorithm at the scale of a sector, region, and module. $x$- and $y$-axis labels refer to the node indices.}
\end{figure}

\begin{figure} 
	\centerline{\includegraphics[width=8.6cm]{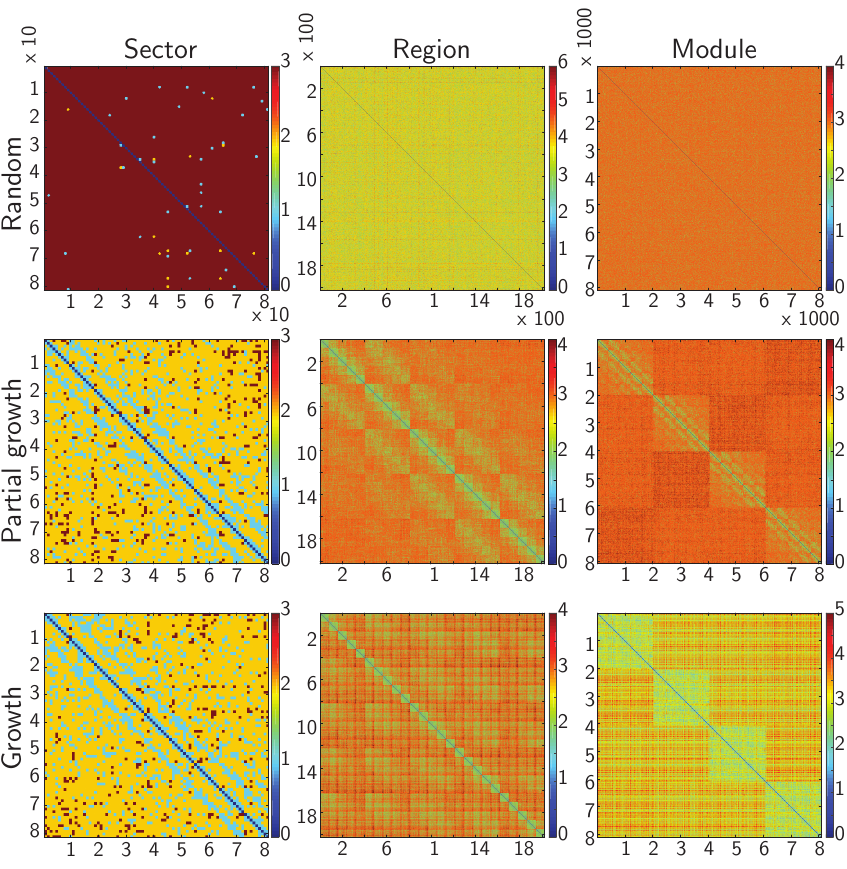}}
	\caption{\label{fig:networks_shortestPaths}Shortest paths of random network, partial growth model, and growth algorithm at the scale of a sector, region, and module. These shortest paths were calculated by first constructing the adjacency matrix of the entire module, and then calculating the shortest paths based on the sub-matrices of the adjacency matrix corresponding to sector, region, and module. Because the random graph has many fewer intra-sector and intra-regional connections, it has larger path lengths on those scales.}
\end{figure}

\begin{figure} 
	\centerline{\includegraphics[width=8.6cm]{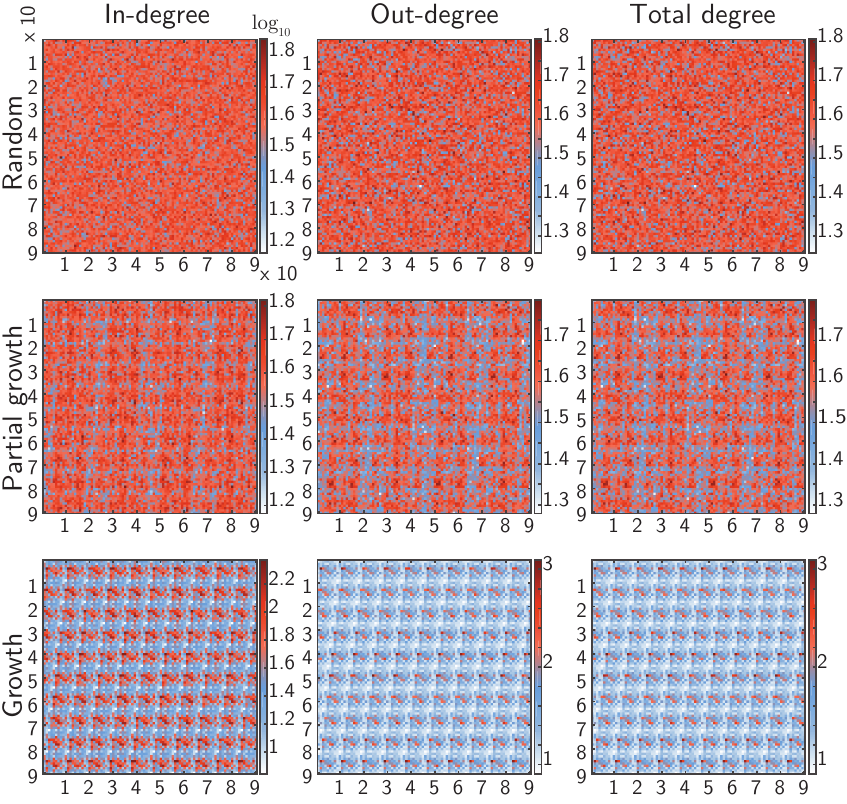}}
	\caption{\label{fig:networks_degreeMatrices}In-degree, out-degree, and total degree of random matrix, partial growth, and growth algorithm. The network constructed with the growth model has a much broader degree distribution as well as more spatial order. Here the $x$- and $y$-axis labels refer to the spatial indices of the nodes in the network.}
\end{figure}

Here we compare several graph metrics of the random network, the partial growth network, and the network created with the growth algorithm. Each network has 8100 nodes, so the maximum possible number of edges of the directed graph is 65,601,900. The random and growth networks have 330,430 edges for a sparsity of 0.50 \%. The partial growth network has 304,365 edges for a sparsity of 0.46 \%. The average total degree of a node in the random network or the growth network is 81.6. In the partial growth network the average total degree is 75.2. For the random network, the minimum total degree is 37 and the maximum total degree is 118. For the partial growth network the minimum total degree is 41 and the maximum total degree is 115. For the growth model the minimum total degree is 17, and the maximum total degree is 1473. Figure \ref{fig:networks_adjacencyMatrices} shows the adjacency matrices of the three networks at the scale of a sector, a region, and the entire module. Figure \ref{fig:networks_shortestPaths} shows the shortest path lengths for the three networks at the scale of a sector, a region, and the entire module. Figure \ref{fig:networks_degreeMatrices} shows the base-ten logarithm of the in-degree, out-degree, and the total degree for the three networks.

\section{\label{apx:spatialPowerLawRamifications}Ramifications of power law degree distribution}
\begin{figure} 
	\centerline{\includegraphics[width=8.6cm]{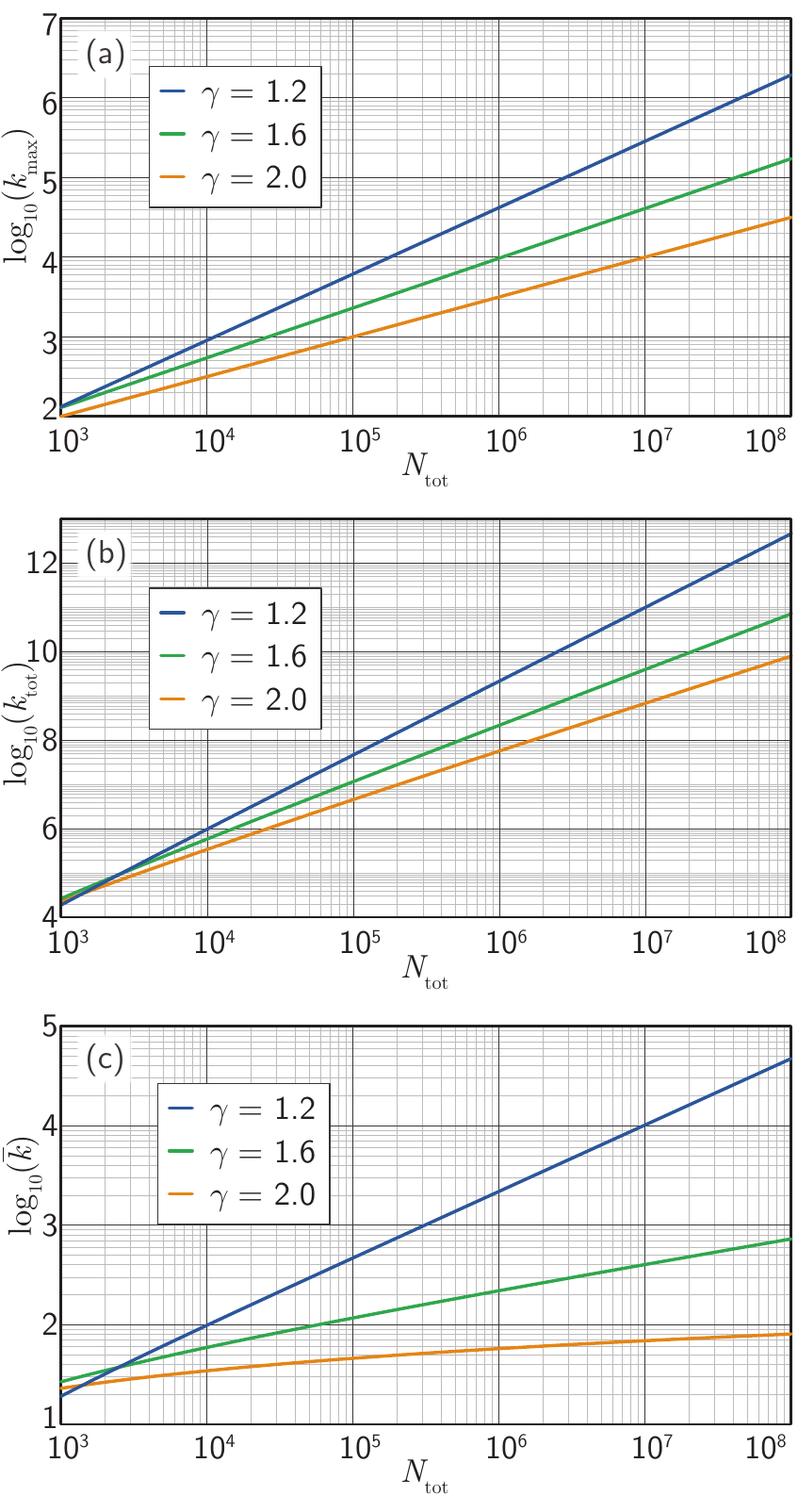}}
	\caption{\label{fig:networks_generalDegreeAnalysis}General degree analysis. (a) Total number of edges in a network as a function of the total number of nodes in the network. (b) Maximum degree of a single node in a network as a function of the total number of nodes in the network. (c) Average degree of a node in the network as a function of total number of nodes in the network.}
\end{figure}
Constraining ourselves to networks with power-law degree distribution allows us to draw general conclusions about the networks we intend to employ. A general statement regarding such networks is that the probability of a node having degree between $k$ and $k+dk$ is given by
\begin{equation}
\label{eq:pOfk}
p(k) = Bk^{-\gamma}.
\end{equation}
We seek expressions for the maximum degree, the total number of edges, and the average degree of nodes in a network with $N_{\mathrm{tot}}$ total nodes.

We assume the probability of a node having degree between $k$ and $k+dk$ is given by Eq.\,\ref{eq:pOfk}. We assume the distribution is valid over a range from $k_{\mathrm{min}}$ to $k_{\mathrm{max}}$. The normalization constant $B = (\gamma-1)/(k_{\mathrm{min}}^{1-\gamma}-k_{\mathrm{max}}^{1-\gamma})$. In a system of $N_{\mathrm{tot}}$ total nodes, we can interpret $N(k) = N_{\mathrm{tot}}p(k)$ as the number of nodes with degree between $k$ and $k+dk$. We can define the maximum degree of the ensemble by $N(k_{\mathrm{max}}) = 1$. For a general network of $N_{\mathrm{tot}}$ neurons characterized by an exponent $\gamma$, we can expect to find zero nodes above $k_{\mathrm{max}}$. Thus, $k_{\mathrm{max}} = [B(k_{\mathrm{max}})N_{\mathrm{tot}}]^{1/\gamma}$. This expression can be used to plot $k_{\mathrm{max}}$ versus $N_{\mathrm{tot}}$, and the result is shown in Fig. \ref{fig:networks_generalDegreeAnalysis}(a) for three values of $\gamma$.

The total number of edges is given by
\begin{equation}
\label{eq:kTot}
k_{\mathrm{tot}} = \int_{k_{\mathrm{min}}}^{k_{\mathrm{max}}}N(k)kdk = N_{\mathrm{tot}}\int_{k_{\mathrm{min}}}^{k_{\mathrm{max}}}kp(k)dk.
\end{equation}
This function is shown in Fig. \ref{fig:networks_generalDegreeAnalysis}(b) for three values of $\gamma$. The average number of edges per node is then given by $\bar{k} = k_{\mathrm{tot}}/N_{\mathrm{tot}}$. This function is plotted in Fig. \ref{fig:networks_generalDegreeAnalysis}(c).

The conclusions drawn from this general analysis are significant for informing decisions about neural hardware. For example, if we are limited to an ensemble of neurons with $N_{\mathrm{tot}} = 10^6$ due to area limitations, we know that 10,000 will be the highest degree of any node if the degree distribution is governed by Eq. \ref{eq:pOfk} with $\gamma = 1.6$. We expect that the highest degree node of the system will represent a limit on information integration. Thus, a limit on the number of neurons in a module translates into a limitation on the information integration of the module. Inversely, if a neural hardware platform is limited at the device level to a small $k_{\mathrm{max}}$, this places a limit on the total number of neurons that can be integrated with power-law connectivity. For example, if a type of neuron cannot achieve greater than 100 synaptic connections, the total number of neurons in the ensemble is limited to 1000. 

\section{\label{apx:sectorRouting}Waveguide routing}
\begin{figure}[h!] 
	\centerline{\includegraphics[width=8.6cm]{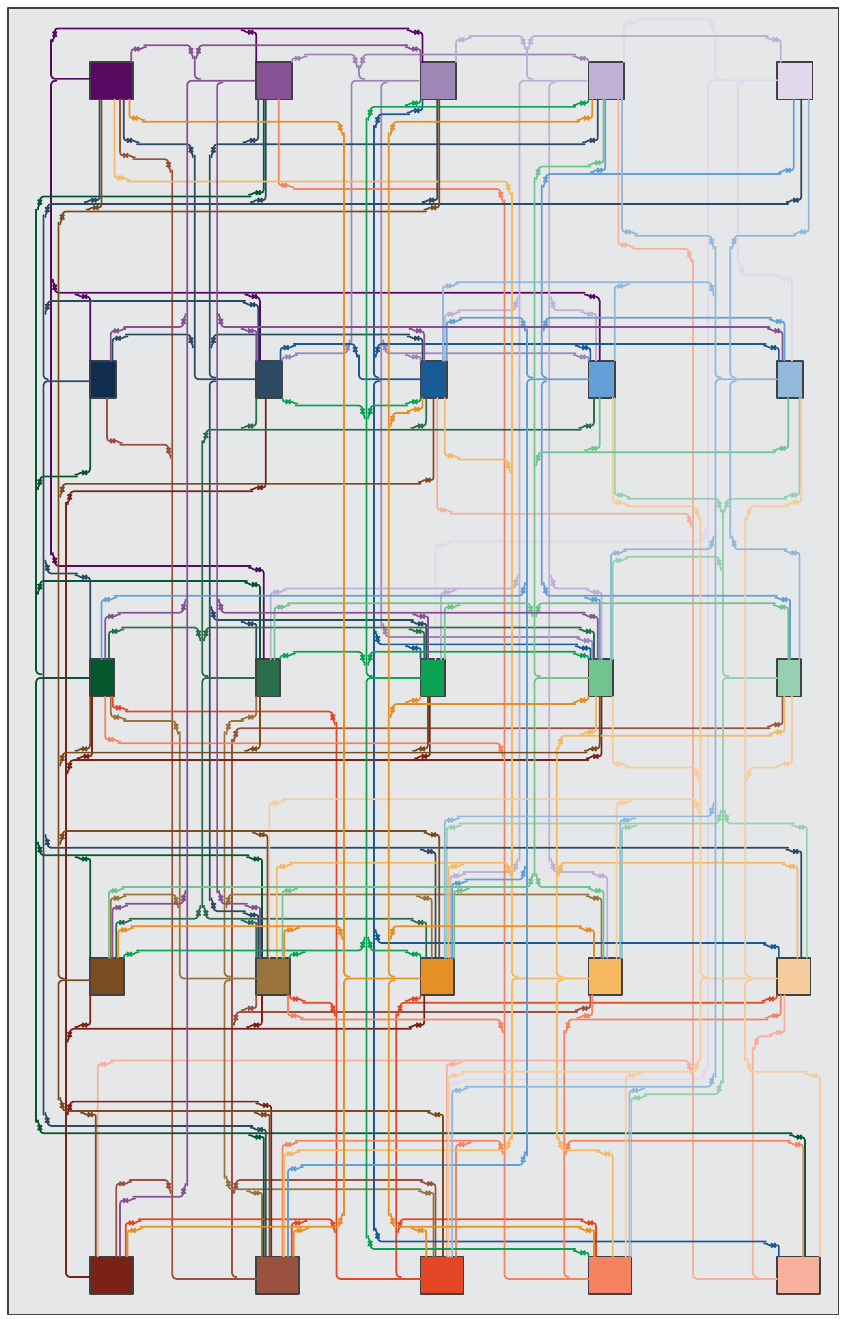}}
	\caption{\label{fig:networks_routingDiagram_full_growth}Row-column routing architecture in 5 $\times$ 5 sector. Growth algorithm.}
\end{figure}
\begin{figure}[t] 
	\centerline{\includegraphics[width=8.6cm]{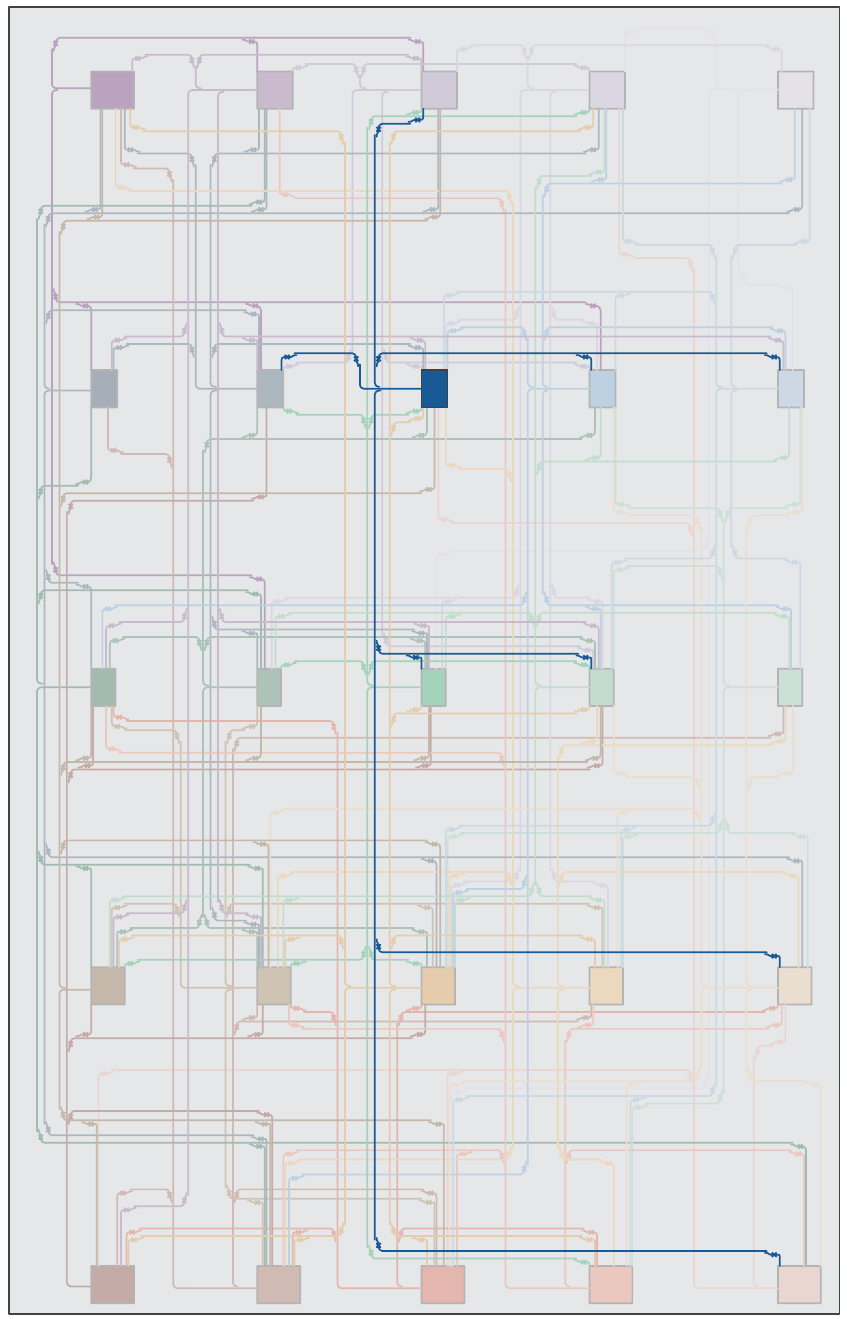}}
	\caption{\label{fig:networks_routingDiagram_full_oneToFew}Row-column routing architecture in 5 $\times$ 5 sector. Growth algorithm. Here we highlight the out-directed connections from a single node.}
\end{figure}
Minimization of wiring is an NP-complete problem, and we make no attempt to solve it in the present context with the constraints particular to SOEN hardware. Instead we develop a practical all-to-all routing scheme and arrive at wiring diagrams by pruning the complete set of connections down to those corresponding to the adjacency matrix of the network under consideration. 

To route the waveguides connecting each neuron to all other neurons within a local sector, we begin with the center node. Light is assumed to exit the node to the west, and it branches into north-running and south-running waveguides on the upper waveguide plane. Each of these waveguides splits east and west at each row to send light to the neurons in that row. Inter-planar couplers \cite{chbu2017} are shown as pairs of triangles in Fig. \ref{fig:networks_routingDiagram_full_allToAll}. Using these couplers, light is routed to the lower plane for east-west routing. It is routed back to the upper plane before entering the SPDs at each neuron. Interplanar transistions are required to mitigate losses due to waveguide crossings. Although intra-planar crossings can be used sparingly, their loss is high enough that they cannot be the exclusive means of crossing waveguides \cite{lish2014}. 

Connections from the central node to all other nodes are established in this manner. Next, a similar approach is taken for the node just north of the central node. Within the central routing column, each subsequent node has two waveguides running north-south. One is to the east of the center node's main north-south waveguide, and one is to the west. Such an approach adds a factor of two to the width of columns, but avoids intra-plane crossings. A better solution is likely possible. At each point where a waveguide must branch, a power tap evanescently couples the appropriate fraction of light to couple the optical signal to all destinations. The objective is to couple as close to one photon as possible to each neuron while avoiding intolerable communication errors. In the power calculation presented in Sec. \ref{sec:physicalInstantiation}, we assume ten photons are produced for each synaptic connection, and reception of one or more photons will trigger a synaptic firing event. 
\begin{figure}
	\centerline{\includegraphics[width=8.6cm]{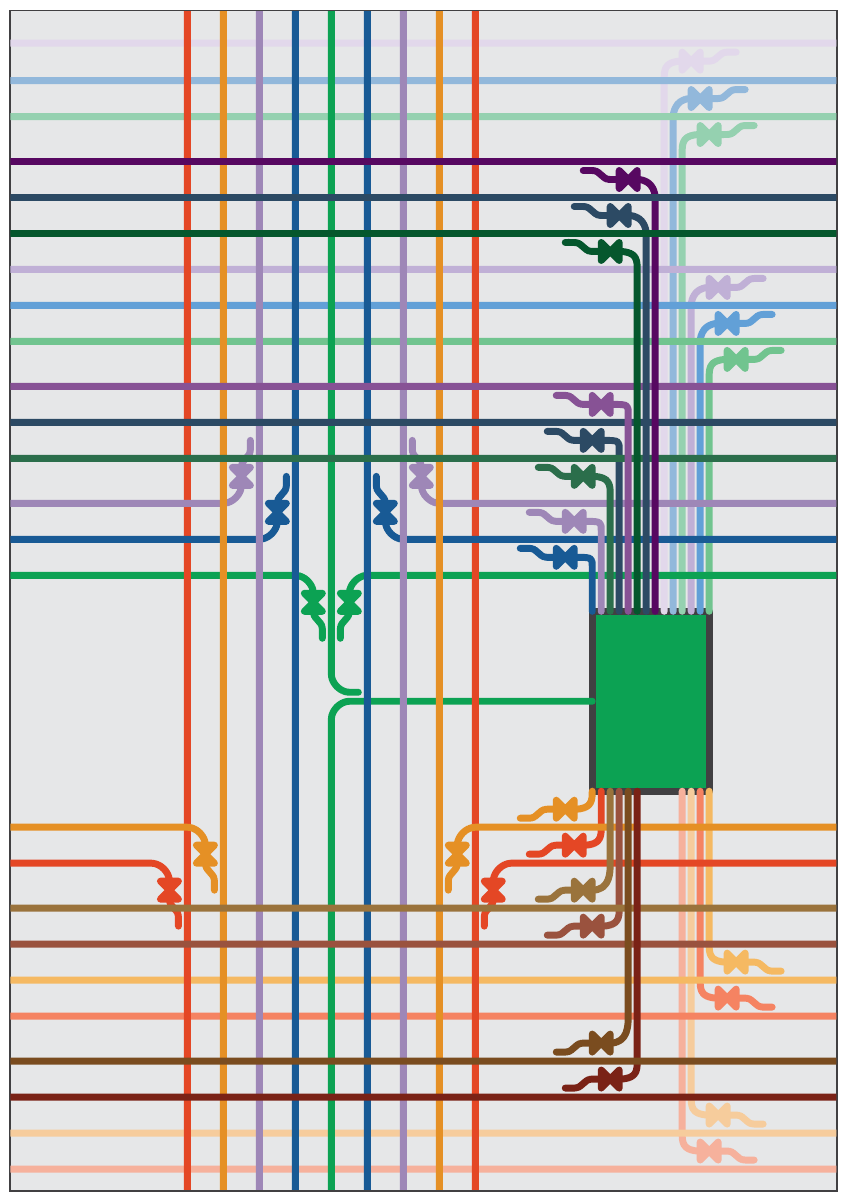}}
	\caption{\label{fig:networks_routingDiagram_zoom_allToAll}Row-column routing architecture in 5 $\times$ 5 sector. All-to-all connectivity. Here we show a spatial zoom near a single node.}
\end{figure}
\begin{figure}
	\centerline{\includegraphics[width=8.6cm]{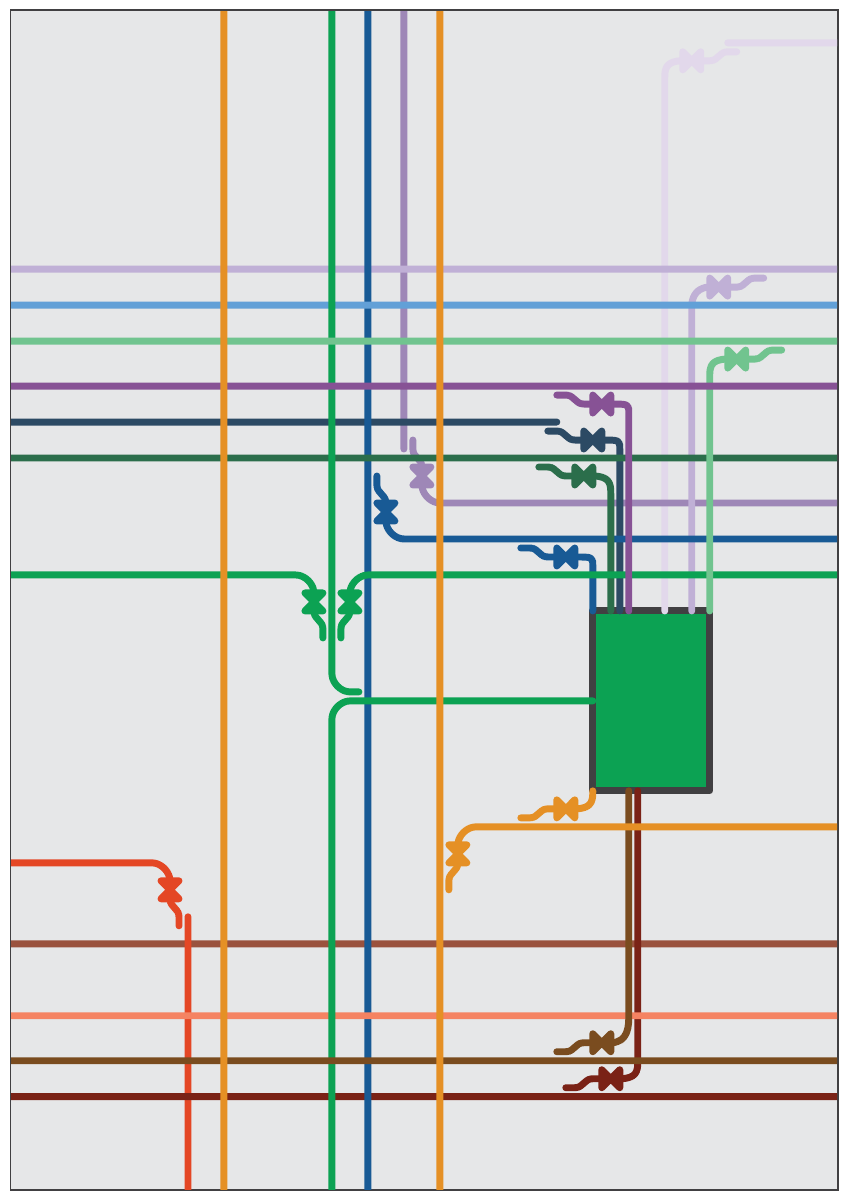}}
	\caption{\label{fig:networks_routingDiagram_zoom_growth}Row-column routing architecture in 5 $\times$ 5 sector. Growth algorithm. Here we show a spatial zoom near a single node.}
\end{figure}

Routing proceeds by establishing the connections from each node up the center column followed by connections from each node below the center node down to the bottom of the center column. As each new row or column waveguide is added, it is placed outside the existing waveguides, leading to dense ribbons of waveguides in the rows and columns, as shown in Fig. \ref{fig:networks_routingDiagram_full_allToAll}. To calculate the width of a column, we define the term
\begin{equation}
\label{eq:a}
\begin{split}
a & = w_{\mathrm{wg}}+g_{\mathrm{tap}}+h_{\mathrm{sine}} \\
& +(w_{\mathrm{wg}}+w_{\mathrm{IPC}})/2.
\end{split}
\end{equation}
Within this model, we calculate the width of a routing column as
\begin{equation}
\label{eq:wCol}
\begin{split}
w_{\mathrm{col}} &= 2 n_{\mathrm{row}}(a+g_{\mathrm{wg}})+2r_{\mathrm{bend}},
\end{split}
\end{equation}
and the height of a routing row as
\begin{equation}
\label{eq:hRow}
\begin{split}
h_{\mathrm{row}} &= n_{\mathrm{N}}(a+n_{\mathrm{spd}}(w_{\mathrm{wg}}+g_{\mathrm{wg}}))+2r_{\mathrm{bend}}.
\end{split}
\end{equation}
In these expressions, $n_{\mathrm{row}}$ is the number of rows in a sector, $g_{\mathrm{tap}}$ is the gap of an evanescent tap, $h_{\mathrm{sine}}$ is the height of sine bend for a power tap, $w_{\mathrm{wg}}$ is the width of a waveguide, $w_{\mathrm{IPC}}$ is the width of an inter-planar coupler, $g_{\mathrm{wg}}$ is the gap between two waveguides sufficient to avoid cross talk, $r_{\mathrm{bend}}$ is the radius of a routing bend, $n_{\mathrm{N}}$ is the number of neurons in a sector, and $n_{\mathrm{spd}}$ is the number of SPDs associated with each synapse. We take $n_{\mathrm{spd}}=3$ in this analysis to account for synaptic firing, Hebbian, and anti-Hebbian synaptic update signals from pre-synaptic neurons. Two SPDs receiving photons from the local, post-synaptic neuron are also required, but they do not need to be accessed via the same routing waveguides, and it is assumed these detectors can be patterned either below or above the routing infrastructure. 

The expressions in Eqs.\,\ref{eq:wCol} and \ref{eq:hRow} approximate the dimensions of routing channels for sectors with all-to-all connectivity, but in practice, connectivity will be much more sparse. In Fig.\,\ref{fig:networks_routingDiagram_full_growth} we show a routing diagram for a 5 $\times$ 5 network designed using the growth algorithm. Connections from a single node are highlighted in Fig. \ref{fig:networks_routingDiagram_full_oneToFew}. Detail of connections to single neurons in the fully connected and growth algorithm cases are shown in Figs. \ref{fig:networks_routingDiagram_zoom_allToAll} and \ref{fig:networks_routingDiagram_zoom_growth}. To account for the sparsity, we multiply the width of Eq.\,\ref{eq:wCol} by $k_{\mathrm{out}}/n_{\mathrm{N}}$ and the height of \ref{eq:hRow} by a factor $k_{\mathrm{in}}/n_{\mathrm{N}}$ when calculating the contribution to the sector area by a neuron with out degree $k_{\mathrm{out}}$ and in degree $k_{\mathrm{in}}$. These in and out degrees refer to the degree calculated using only the adjacency matrix of the sector. That is, only connections which originate and terminate within a sector are accounted for in this part of the analysis. Distant connections are accounted for below.

The expressions in Eqs.\,\ref{eq:wCol} and \ref{eq:hRow} account for the area consumed by routing waveguides, but the contribution to total sector area due to the area of the neurons must also be included. We approximate the size of a neuron with
\begin{equation}
\label{eq:hNeuron}
h_{\mathrm{neuron}} = 2L_{\mathrm{spd}},
\end{equation}
and
\begin{equation}
\label{eq:wNeuron}
\begin{split}
w_{\mathrm{neuron}} &= k_{\mathrm{in}}(w_{\mathrm{wg}}+g_{\mathrm{wg}})\\
&+\frac{3}{2}(L_{\mathrm{tap}}+L_{\mathrm{sine}}+L_{\mathrm{DeMux}}+L_{\mathrm{IPC}}+r_{\mathrm{bend}}).
\end{split}
\end{equation}
In these expressions, $k_{\mathrm{in}}$ is the in-degree of the neuron, $L_{\mathrm{tap}}$ is the length of an average evanescent power tap, $L_{\mathrm{sine}}$ is the length of a sine in a power tap, $L_{\mathrm{DeMux}}$ is the length of a de-multiplexer used to separate the photons for synaptic firing, Hebbian, and anti-Hebbian update. While it is not necessary to use different colors for these different operations, it may be advantageous so the different operations can be turned on and off independently. If multiple colors are used, the de-multiplexer is likely to be a frequency-dependent evanescent coupler. $L_{\mathrm{IPC}}$ is the length of an inter-planar coupler.

To calculate the area of a sector, the areas occupied by routing and by neurons are summed, and the result is divided by the number of pairs of waveguide planes used in the fabrication process. Once the area of a sector has been calculated, we need to consider the area consumed by routing between sectors to form regions and between regions to form a module. For these calculations, the expression of Eqs.\,\ref{eq:wCol} and \ref{eq:hRow} are used, except $n_{\mathrm{row}}$ refers to the number of rows of neurons in the region (which is the number in a sector times the number of rows of sectors in the region), and $n_{\mathrm{N}}$ refers to the number of neurons in the region. The in and out degrees used now refer to connections originating in one sector of the region and terminating in a different sector of the same region. When calculating the area of a region, the area of the neurons is not added, as all inter-sector synapses are assumed to be above the intra-sector synapses on higher waveguide planes. This simplification can be made if the number of synapses resulting from intra-sector connections is equal to or less than intra-regional and intra-modular connections. This corresponds to a Rentian exponent less than one. For the case of the growth model network with 8100 neurons, the average number of intra-sector connections is 17.1. The average number of intra-regional connections is 17.8. The average number of intra-modular connections is 17.6. For the random network, these numbers are 0.4, 9.8, and 91.9, respectively. Routing and area analysis of the random network requires a different approach.

For the network of 8100 neurons considered in Sec.\,\ref{sec:networkConstruction}, there are three levels of hierarchy: sectors, regions, and the total module. The analysis of the area of routing waveguides in the module is nearly identical to the analysis of the regions with the appropriate substitutions of $n_{\mathrm{row}}$, $n_{\mathrm{N}}$, $k_{\mathrm{in}}$, and $k_{\mathrm{out}}$.

The size of the neurons and routing depend on the number of connections made by each neuron. Using the above expressions for the contribution of a neuron to the total area, we can calculate $A_n(k)$, the area of a neuron with degree $k$. For the specific case of the network of 8100 neurons, we can explicitly calculate this area for each neuron because the adjacency matrix is known. In a general case, we can calculate the total area of a general network if the number of neurons and their degree distribution, $p(k)$, is specified. The total area of the network is given by
\begin{equation}
\label{eq:networkArea}
A_{\mathrm{N}} = N_{\mathrm{tot}}\int_{k_{\mathrm{min}}}^{k_{\mathrm{max}}}A_n(k)p(k)dk.
\end{equation}
For the analysis presented in Fig.\,\ref{fig:networks_scaling}(a), we use this expression to calculate the total network area, and we obtain $A_n(k)$ by fitting the data from the network of 8100 neurons to a power law. From this fit, we find $A_n(k)\propto k^{1.4}$. To generate Fig.\,\ref{fig:networks_scaling}(a), we assume the in-degree and out-degree are the same. 

The following parameters have been used in these calculations. The width of a waveguide, $w_{\mathrm{wg}} = 500$\,nm. The gap between waveguides, $g_{\mathrm{wg}} = 1$\,\textmu m. The height of a sine bend, $h_{\mathrm{sine}} = 1$\,\textmu m. The length of a sine bend is the same. The gap of an evanescent power tap, $g_{\mathrm{tap}}= 500$ nm. The length of a power tap, $L_{\mathrm{tap}} = 5$\,\textmu m. The length of an inter-planar coupler, $L_{\mathrm{IPC}} = 36$\,\textmu m. The width of an inter-planar coupler, $w_{\mathrm{IPC}} = 4$\,\textmu m. The length of an SPD, $L_{\mathrm{SPD}} = 10$\,\textmu m. The radius of a routing bend, $r_{\mathrm{bend}} = 2$\,\textmu m. The length of a de-multiplexer, $L_{\mathrm{DeMux}} = 5$\,\textmu m.
These parameters are intended to model silicon passive photonic components. For longer-distance connections, silicon nitride waveguides may be preferable due to the lower index contrast and therefore lower loss. The area of silicon nitride passive waveguides is roughly twice that of silicon.  

\begin{figure} 
	\centerline{\includegraphics[width=8.6cm]{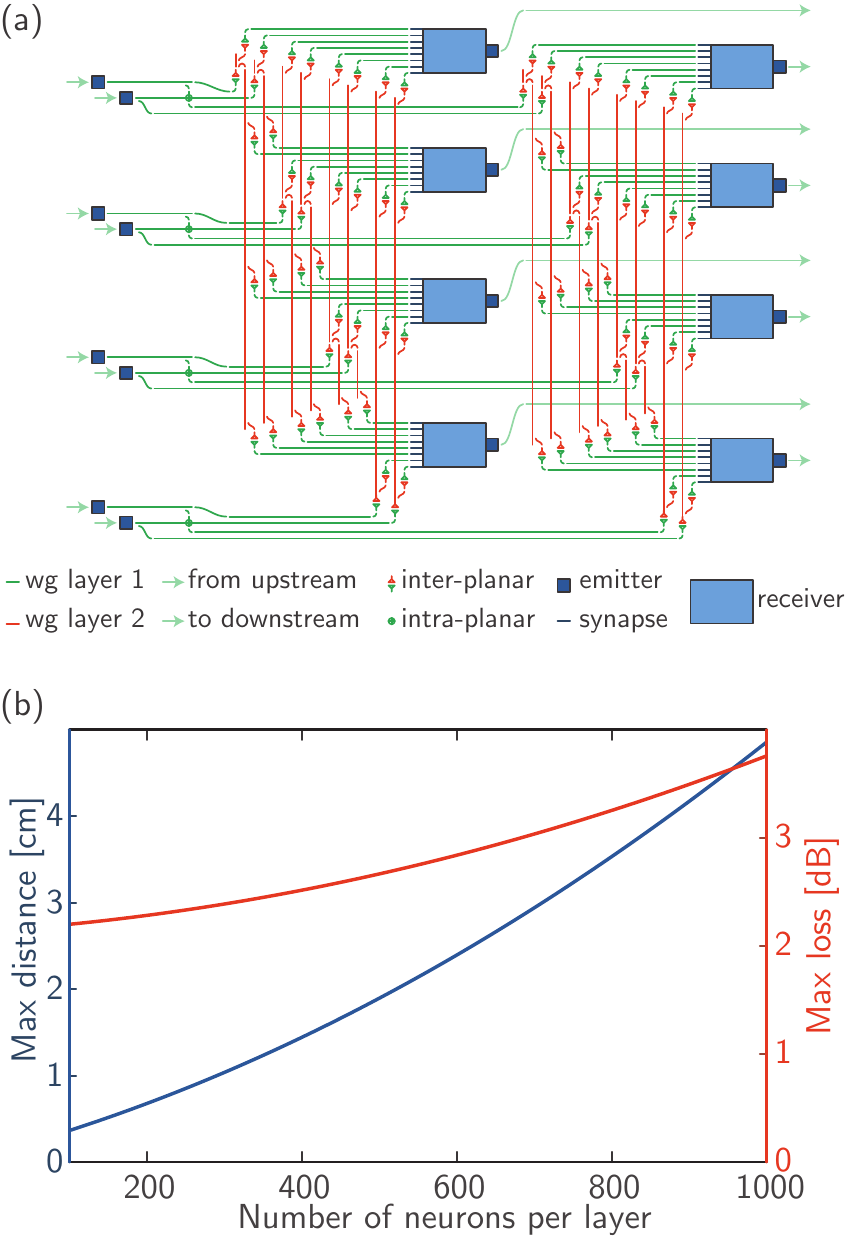}}
	\caption{\label{fig:networks_feedForward}Routing for a feed-forward network. (a) All possible connections from upper left node. (b) All possible connections from middle node.  Routing diagram for a $5 \times 5$ network designed with the growth algorithm.}
\end{figure}

While the sector comprising a square array of nodes shown in Fig.\,\ref{fig:networks_routingDiagram_full_allToAll} works well for the growth algorithm leading to power-law degree distribution, many applications leverage feed-forward networks that are very wide with only a few layers but all-to-all connectivity from one layer to the next. A routing diagram for such a network is shown in Fig.\,\ref{fig:networks_feedForward}.
Here we again use pairs of waveguide planes for north-south and east-west routing. Figure \ref{fig:networks_feedForward}(b) shows the spatial scaling and loss as a function of the number of neurons in a network layer for a process employing 12 waveguide planes to form two layers of a feed-forward network. The maximum distance refers to the row-column propagation distance from a neuron at the bottom of one network layer to the top of an adjacent layer. It is assumed waveguides have 0.2 dB/cm propagation loss \cite{ch2018}. For 1000 neurons per layer, the system would fit on a chip 2.5\,cm $\times$ 2.5\,m.

\section{\label{apx:manufacturing}Approaches to lithography}
Our emphasis on comparison to the size of a 300\,mm wafer is due to the requirement of manufacturability. There are multiple considerations regarding the fabrication of these devices with existing 300-mm lithographic infrastructure. For example, in the construction of the network adjacency matrix described in Sec.\,\ref{sec:networkConstruction}, each time we expand our network to a new level of hierarchy, we begin by tiling the adjacency matrix of the lower level of hierarchy along the diagonal. This provides a scalable route to fabrication and an economical reticle utilization. Fabrication across a wafer would begin by first patterning all the identical local sectors with dedicated planes of waveguides routing only intra-sector connections. During the next step of fabrication, regions comprising multiple sectors would be patterned with a larger reticle, but again all regions would be identical at this processing layer, as the waveguide planes patterned at this step would be dedicated exclusively for intra-regional connectivity. The same approach would be taken to scale regions to modules. With each vertically stacked waveguide patterning, we must also deposit and pattern layers of superconductors including single-photon detectors and inductors, and Josephson junctions, though junctions may require fewer layers. 

The challenge for this hierarchical approach to fabrication will be that ultimately a reticle image spanning the  wafer is desirable to form the highest level of hierarchy. This is not presently possible in 300\,mm lithography. It may be possible to pattern the highest layers by stitching fields. Each field would be unique, so reticle sets would be expensive. Alternatively, for the final layers of processing, direct-write lithography may be more advantageous. 

While identical tiling within layers of hierarchy is most advantageous in design and in use of reticles, it is not clear what effects it will have on the computational capacity of the network. To achieve differentiated processing we need unique processors. Sectors made from the same reticles would be structurally different at higher levels of hierarchy. Different synaptic states would also achieve differentiation dynamically. And different sectors could receive different sensory stimuli. The structural requirements for functional differentiation and integration remains an open field of research.

\section{\label{apx:power}Power consumption}
\begin{figure} 
	\centerline{\includegraphics[width=8.6cm]{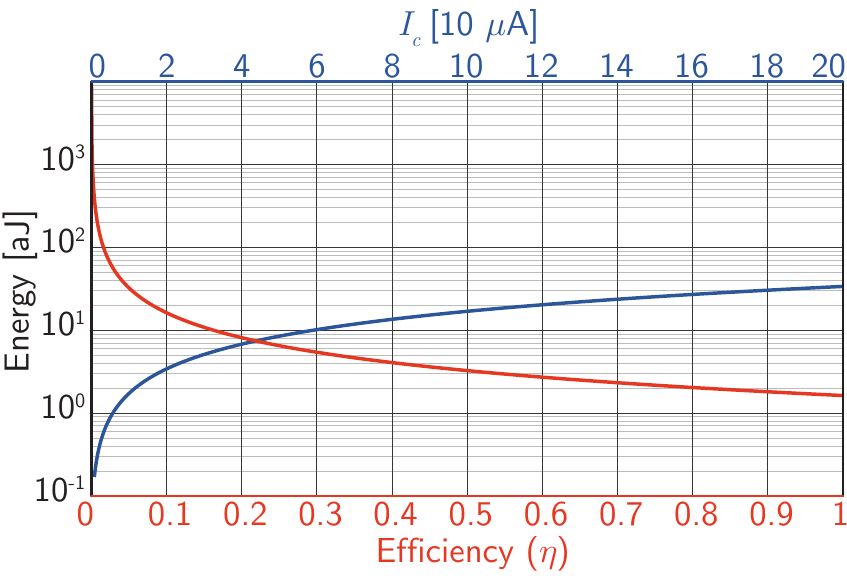}}
	\caption{\label{fig:networks_contributionsToEnergy}Photonic and electronic contributions to power. The red trace is the term $\zeta h \nu/\eta$, and the blue trace is the term $\chi n_{\mathrm{fq}} I_c \Phi_0$.}
\end{figure}

For this analysis we model the energy to produce a photon as $E = h\nu/\eta$, where $h$ is Planck's constant, $\nu$ is the frequency of light (250\,THz if the emitters of Ref.\,\onlinecite{buch2017} are employed), and $\eta$ is an efficiency factor that takes into account all energy used by the thresholding junction, the superconducting amplifiers, and the light emitter. From Ref.\,\onlinecite{sh2018d} we take $\eta = 10^{-4}$. The energy of a neuronal firing event is given by $E_{\mathrm{out}} = \zeta h \nu k_{\mathrm{out}}/\eta $, where $\zeta$ is the number of photons produced for each synaptic connection. In principle, this can be as low as one, but such low photon numbers would make communication noisy and unreliable. $\zeta = 10$ is likely to be a safe operating point \cite{sh2018d}. To calculate the total energy per neuronal firing event, we also need to sum the energy of all the synaptic firing events that led to the neuronal firing event. This energy is given by $E_{\mathrm{in}} = \chi n_{\mathrm{fq}} I_c \Phi_0 k_{\mathrm{in}}$ where $n_{\mathrm{fq}}$ is the number of fluxons generated in a synaptic firing event (taken to be 245), and $\chi$ is the fraction of synapses that must fire to drive the neuron to threshold. We assume $\chi = 1/3$. To simplify analysis, let us take $k_{\mathrm{in}} = k_{\mathrm{out}} = k$. The total energy of a neuronal firing event is thus
\begin{equation}
\label{eq:neuronalFiringEvent}
E(k) = (\zeta h \nu/\eta + \chi n_{\mathrm{fq}} I_c \Phi_0) k\equiv E_0 k.
\end{equation}
The two contributions to Eq.\,\ref{eq:neuronalFiringEvent} are due to generation of photons during a neuronal firing event ($\zeta h \nu/\eta$), and generation of fluxons during a synaptic firing event ($\chi n_{\mathrm{fq}} I_c \Phi_0$). These two terms are plotted independently in Fig.\,\ref{fig:networks_contributionsToEnergy}. We see the photon production efficiency, $\eta$, must be very high for the photonic contribution to be less than the fluxonic contribution. Based on this analysis, we conclude that the choice to use junctions with 10\,\textmu A $I_c$ for the synaptic receivers of Ref.\,\onlinecite{sh2018b} was probably misguided, and it may be the case that all the circuits of Refs.\,\onlinecite{sh2018b} and \onlinecite{sh2018c} can be produced with junctions of 40\,\textmu A $I_c$. 

The power dissipated by a neuron is given by the energy per neural firing event multiplied by the number of neural firing events per unit time, 
\begin{equation}
\label{eq:neuronPower}
P_{\mathrm{n}}(k,f) = E(k)f, 
\end{equation}
where the subscript $\mathrm{n}$ denotes the power of a single neuron as opposed to the total power of the network. 

We assume the network is characterized by a power law degree distribution of the form
\begin{equation}
\label{eq:degreeProbability}
p(k) = B_1 k^{-\gamma},
\end{equation} 
meaning the probability that a node from the network chosen at random will have degree between $k$ and $k+dk$ is $p(k)$. We further assume the network is characterized by a power law frequency distribution of the form
\begin{equation}
\label{eq:frequencyProbability}
p(f) = B_2 f^{-\mu},
\end{equation}
meaning that, during a period of observation, a node chosen at random from the network will be observed to spike with frequency between $f$ and $f+df$ is given by $p(f)$. Incorporating these probability distributions, and assuming there are $N_{\mathrm{tot}}$ nodes in the network, the total network power consumption is given by
\begin{equation}
\label{eq:networkAverageTotalPower}
P_{\mathrm{N}} = N_{\mathrm{tot}}\int_{f_{\mathrm{min}}}^{f_{\mathrm{max}}} \int_{k_{\mathrm{min}}}^{k_{\mathrm{max}}} P_{\mathrm{n}}(k,f)p(k)p(f) df dk, 
\end{equation}
and the power density is $P_{\mathrm{N}}/A_{\mathrm{N}}$, where $A_{\mathrm{N}}$ is the total area of the network.

Considering first the spatial term of Eq.\,\ref{eq:networkAverageTotalPower}, we see that 
\begin{equation}
\label{eq:networkPower_spatialConsideration1}
P_{\mathrm{N}} \propto \int_{k_{\mathrm{min}}}^{k_{\mathrm{max}}} k^{-(\gamma-1)}dk, 
\end{equation}
whereas
\begin{equation}
\label{eq:networkPower_spatialConsideration2}
A_{\mathrm{N}} \propto \int_{k_{\mathrm{min}}}^{k_{\mathrm{max}}} k^{-(\gamma-1.4)}dk. 
\end{equation}
The exponent of 1.4 is explained in Appendix \ref{apx:sectorRouting}. Because the integrand of $A_{\mathrm{N}}$ decays more slowly with $k$ than the integrand of $P_{\mathrm{N}}$, power density will decrease as a function of network size, and will therefore never limit scaling. The exponents $\gamma$ and $\mu$ can also be chosen to adjust power density.

Considering the frequency term of Eq.\,\ref{eq:networkAverageTotalPower}, we see that 
\begin{equation}
\label{eq:networkPower_frequencyConsideration1}
P_{\mathrm{N}} \propto \int_{f_{\mathrm{min}}}^{f_{\mathrm{max}}} f^{-(\mu-1)}df. 
\end{equation}
The integrand in Eq.\,\ref{eq:networkPower_frequencyConsideration1} is the power spectral density of the network, 
\begin{equation}
\label{eq:networkPower_frequencyConsideration2}
P_{\mathrm{N}}(f) \propto f^{-(\mu-1)}. 
\end{equation}
If the exponent in the frequency power law distribution (Eq.\,\ref{eq:frequencyProbability}) is $\mu = 2$, the network-averaged power spectrum of Eq. \ref{eq:networkPower_frequencyConsideration2} takes the $1/f$ form observed in cortical networks \cite{budr2004}, believed to be significant for cognition \cite{bu2006}, and known to be related to self-organized criticality and fractal objects \cite{be2007,bata1987,yara2017}. In the hardware platform under consideration, $\mu$ can be adjusted through the threshold bias currents as well as the light-emitter gain bias currents, thus providing a means to investigate network operation as a function of power spectral density.

Using Eq.\,\ref{eq:networkAverageTotalPower} to calculate the power consumption of the network of 8100 neurons, and assuming $f_{\mathrm{min}} = 100$\,Hz and $f_{\mathrm{max}} = 20$\,MHz, we find the network will consume 2\,mW. A network of 1 million neurons on a 300\,mm wafer will consume 2\,W. To calculate these numbers, we have assumed $\gamma = 1.4$ and $\mu  = 2$.

\bibliography{bibliography_modelingSOENs}

\end{document}